\documentclass[runningheads]{llncs}
\usepackage{graphicx}
\usepackage{comment}
\usepackage{amsmath,amssymb} % define this before the line numbering.
\usepackage{color}

\usepackage{times}
\usepackage{epsfig}
\usepackage{graphicx}
\usepackage{amsmath}
\usepackage{amssymb}
\usepackage{lib}
\usepackage{multirow}
\usepackage{float}
\usepackage{comment}
\usepackage{xcolor,colortbl}
\usepackage{makecell}
\usepackage[]{algorithm2e}
\usepackage{enumitem}

\newcommand{\OURS}{Mask2CAD}

\newlength\savewidth\newcommand\shline{\noalign{\global\savewidth\arrayrulewidth
  \global\arrayrulewidth 1pt}\hline\noalign{\global\arrayrulewidth\savewidth}}

\newcommand{\tablestyle}[2]{\setlength{\tabcolsep}{#1}\renewcommand{\arraystretch}{#2}\centering\footnotesize}

\definecolor{Gray}{gray}{0.95}
\definecolor{Gray2}{rgb}{0.4, 0.4, 0.4}
\definecolor{LightCyan}{rgb}{0.88,1,1}
\definecolor{Lavender}{rgb}{0.909, 0.909, 0.95}

\newcolumntype{a}{>{\columncolor{Lavender}}c}
\newcolumntype{b}{>{\columncolor{white}}c}
\newcolumntype{T}{>{\footnotesize}c} % define a new column type for \tiny

\usepackage{mathtools}

\begin{document}
% \linenumbers
\pagestyle{headings}
\mainmatter
\def\ECCVSubNumber{193}  % Insert your submission number here

\title{
\OURS: 3D Shape Prediction by Learning to Segment and Retrieve
}

%\end{comment}
%******************

% \begin{comment}
\titlerunning{\OURS}
\author{Weicheng Kuo\inst{1,2} \and
Anelia Angelova\inst{1,2} \and
Tsung-Yi Lin\inst{1,2} \and
Angela Dai\inst{3}}
\institute{Google AI \and Robotics at Google \and Technical University of Munich \\ \email{\{weicheng,anelia,tsungyi\}@google.com, angela.dai@tum.de}}
\authorrunning{W. Kuo et al.}

% \end{comment}
%******************
\maketitle

\begin{abstract}

Object recognition has seen significant progress in the image domain, with focus primarily on 2D perception. 
We propose to leverage existing large-scale datasets of 3D models to understand the underlying 3D structure of objects seen in an image by constructing a CAD-based representation of the objects and their poses.
We present \OURS{}, which jointly detects objects in real-world images and for each detected object, optimizes for the most similar CAD model and its pose.
We construct a joint embedding space between the detected regions of an image corresponding to an object and 3D CAD models, enabling retrieval of CAD models for an input RGB image. This produces a clean, lightweight representation of the objects in an image; this CAD-based representation ensures a valid, efficient shape representation for applications such as content creation or interactive scenarios, and makes a step towards understanding the transformation of real-world imagery to a synthetic domain.
Experiments on real-world images from Pix3D demonstrate the advantage of our approach in comparison to state of the art. To facilitate future research, we additionally propose a new image-to-3D baseline on ScanNet which features larger shape diversity, real-world occlusions, and challenging image views. 
\end{abstract}

\vspace{-4mm}
\section{Introduction}
\seclabel{sec:Intro}

Object recognition and localization in images has been a core task of computer vision with a well-studied history.
Recent years have shown incredible progress in identifying objects in RGB images by predicting their bounding boxes or segmentation masks~\cite{duan2019centernet,he2017mask,lin2017focal}.
Although these advances are very promising, recognizing 3D attributes of objects such as shape and pose is crucial to many real-world applications.
In fact, 3D perception is fundamental towards human understanding of imagery and real-world environments -- from a single RGB image a human can easily perceive geometric structure, and is paramount for enabling higher-level scene understanding such as inter-object relationships, or interaction with an environment by exploration or manipulation of objects.

At the same time, we are now seeing a variety of advances in understanding the shape of a single object from image view(s), driven by exploration of various geometric representations: voxels~\cite{choy20163d,tatarchenko2017octree,wu2016learning}, points~\cite{fan2017point,yang2019pointflow}, meshes~\cite{dai2019scan2mesh,groueix2018papier,wang2018pixel2mesh}, and implicit surfaces~\cite{mescheder2019occupancy,park2019deepsdf}.
While these generative approaches have shown significant promise in inferring the geometry of single objects, these approaches tend to generate geometry that may not necessarily represent a valid shape, with tendency towards noise or oversmoothing, and excessive tessellation.
Such limitations render these results unsuitable for many applications, for instance content creation, real-time robotics scenarios, or interaction in mixed reality environments.
In addition, the ability to digitize the objects of real world images to CAD models opens up new possibilities in helping to bridge the real-synthetic domain gap by transforming real-world images to a synthetic representation where far more training data is available.

In contrast, we propose \OURS{} to join together the capabilities of 2D recognition and 3D reconstruction by leveraging CAD model representations of objects.
Such CAD models are now readily available~\cite{shapenet2015} and represent valid real-world object shapes, in a clean, compact representation -- a representation widely used by existing production applications.
Thus, we aim to infer from a single RGB image object detection in the image as well as 3D representations of each detected object as CAD models aligned to the image view.
This provides a geometrically clean, compact reconstruction of the objects in an image, and a lightweight representation for downstream applications.

\begin{figure}
    \centering
    \includegraphics[width=0.9\textwidth]{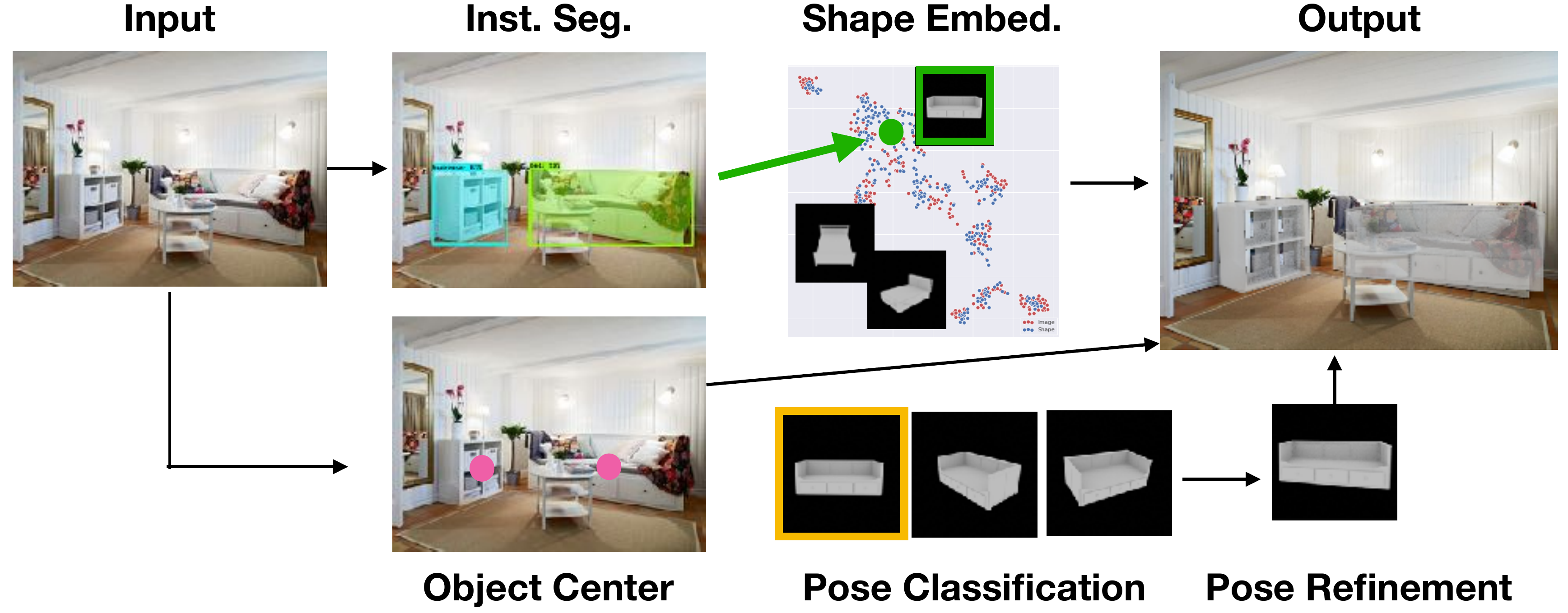}
    \caption{\OURS{} aims to predict object mask, and 3D shape in the scene. We achieve this by formulating an image-shape embedding learning problem. Combined with pose and object center prediction, \OURS{} outputs realistic 3D shapes of objects from a single RGB image input. 
    The entire system is differentiable and learned end-to-end.}
    \label{fig:teaser}
\end{figure}

Our \OURS{} approach jointly detects object regions in an image and learns to map these image regions and CAD models to a shared embedding space (See Figure \ref{fig:teaser}). 
At train time we learn a joint embedding which brings together corresponding image-CAD pairs, and pushes apart other pairs. 
At test time, we retrieve shapes by their renderings from the embedding space. 
To align the shapes to the image, we develop a pose prediction branch to classify and refine the shape alignment.
We train our approach on the Pix3D dataset~\cite{sun2018pix3d}, achieving more accurate reconstructions than state of the art. Importantly, our retrieval-based approach allows adaptation to new domain by simply adding CAD models to the CAD model set without any re-training. Experiments on unseen shapes of the Pix3D dataset~\cite{sun2018pix3d} show notable improvement when we have access to all CAD models at test time (but no access to corresponding RGB images of the unseen models and no re-training). By leveraging CAD models as shape representation, we are able to predict multiple distinct 3D objects per image efficiently (approximately 60ms per image).

In addition to Pix3D, we also apply \OURS{} on ScanNet and propose the first single-image to 3D object reconstruction baseline. Compared to Pix3D, this dataset contains 25K images, an order of magnitude more 3D shapes, complex real-world occlusions, diverse views and lighting conditions. Despite these challenges, \OURS{} still manages to  place appropriate CAD models that match the image observation (see Figure \ref{fig:scannet}). We hope \OURS{} could serve as a benchmark for future retrieval methods and reference for generative methods.

\OURS{} opens up possibilities for object-based 3D understanding of images for content creation and interactive scenarios, and provides an initial step towards transforming real images to a synthetic representation.

\begin{figure}
    \centering
    \includegraphics[width=0.8\textwidth]{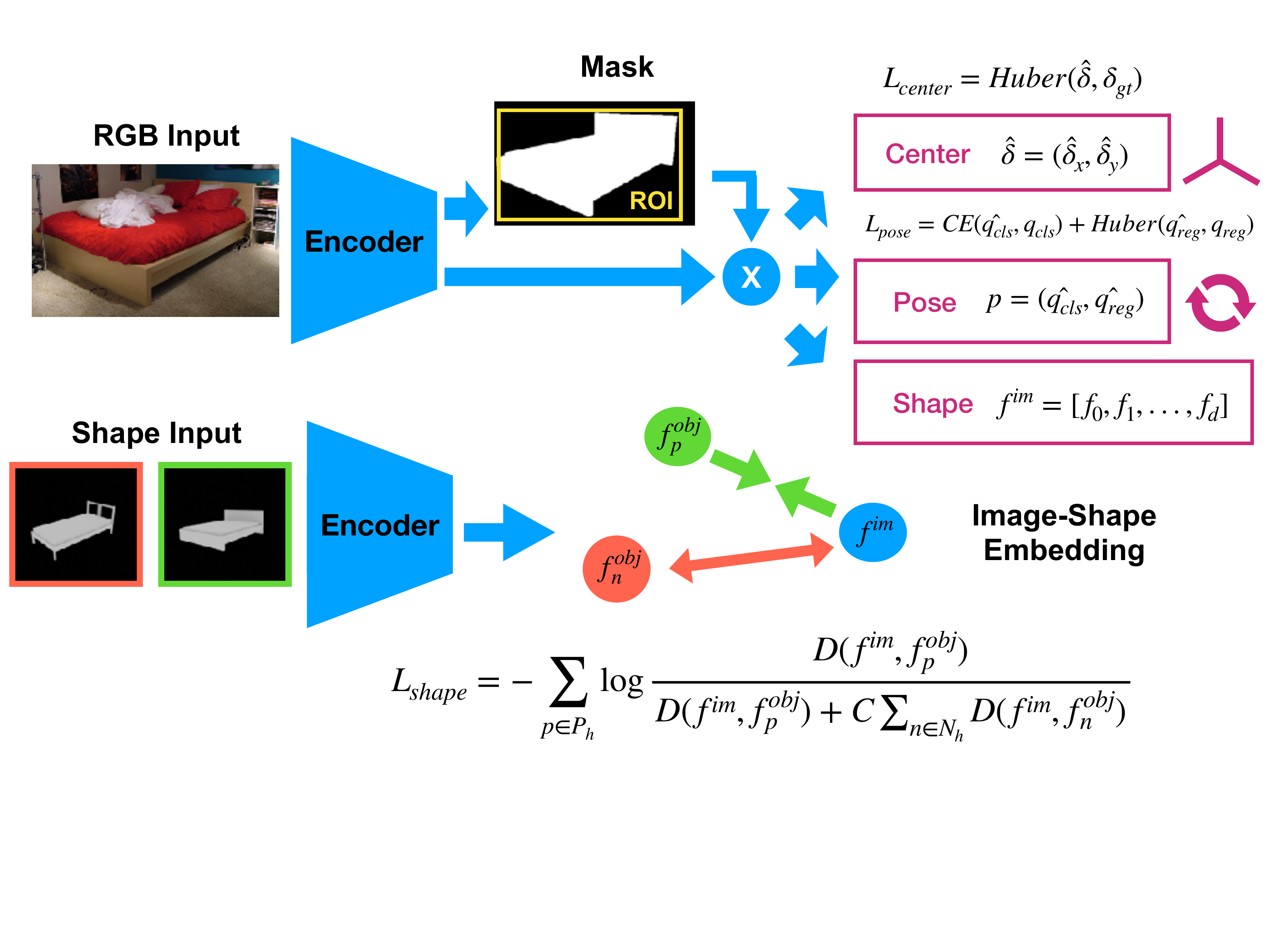}
    \vspace{-15mm}
    \caption{Overview of our \OURS{} approach for joint object segmentation and shape retrieval from a single RGB image.
    At train time, object detection is performed on an RGB image to produce a bounding box, segmentation mask, and feature descriptor for each detected object. 
    The object feature descriptor is then used to train for an image-CAD embedding space for shape retrieval, as well as pose regression for the object rotation and center regression for its location.
    The embedding space is constructed through a triplet loss with corresponding and non-corresponding shapes to a detected object region of an image.
    }
    \label{fig:overview}
\end{figure}
\vspace{-6mm}

\section{Related work}
\label{sec:related}

{\bf Object Recognition in Images.}
Our work draws inspiration from the success of 2D object detection and segmentation in the image domain, where myriad methods have been developed to predict 2D object bounding boxes and class labels from a single image input~\cite{girshick2015fast,lin2017focal,liu2016ssd,redmon2016you,ren2015faster}.
Recent approaches have focused on additionally predicting instance masks for each object~\cite{he2017mask,kuo2019shapemask}.
We build from this 2D object detection and segmentation to jointly learn to predict shape as well.

\medskip
\noindent {\bf Single-View Object Reconstruction.}
In recent years, a variety of approaches have been developed to infer 3D shape from a single RGB image observation, largely focusing on the single object scenario and exploring a variety of shape representations in the context of learning-based methods.
Regular voxel grids are a natural extension of the 2D image domain, and have been shown to effectively predict global shape structures~\cite{choy20163d,wu2016learning}, but remain limited by computational and memory constraints when scaling to high resolutions, as well as uneconomical in densely representing free space.
Thus methods which focus representation power on surface geometry have been developed, including  hierarchical approaches on voxels such as octrees~\cite{riegler2017octnet,tatarchenko2017octree}, or point-based representations~\cite{fan2017point,yang2019pointflow}.
More recently, methods have been developed to predict triangle meshes, largely based on strong topological assumptions such as deforming an existing template mesh~\cite{wang2018pixel2mesh} while free-form generative approaches tend to remain limited by computational complexity to very small numbers of vertices~\cite{dai2019scan2mesh}.
Implicit reconstruction approaches have also shown impressive shape reconstruction results at relatively high resolution by predicting the occupancy~\cite{mescheder2019occupancy} or signed distance field value~\cite{park2019deepsdf} for point sampled locations.

Mesh R-CNN~\cite{meshrcnn} pioneered an approach to extend such single object reconstruction to jointly detect and reconstruct shape geometry for each detected object in an RGB image. 
Mesh R-CNN extends upon the object recognition pipeline of Mask R-CNN~\cite{he2017mask} to predict initial voxel-based occupancy of an object, which is then refined by a graph convolutional network to produce an output mesh for each object.

While these approaches for object reconstruction have shown significant promise in predicting general, structural shape properties, due to the low-level nature of the reconstruction approaches (generating on a per-voxel/per-point basis), the reconstructed objects tend to be noisy or oversmoothed, may not represent valid real-world shapes (e.g., disconnected in thin regions, missing object symmetries), and inefficiently represented in geometry (e.g., over-tessellated to achieve higher resolutions).
In contrast, our \OURS{} approach leverages existing CAD models of objects to jointly segment and retrieve the 3D shape for each object in an image, producing both an accurate reconstruction and clean, compact geometric representation.

\medskip
\noindent {\bf CAD Model Priors for 3D Reconstruction.}
Leveraging geometric model-based priors for visual understanding has been established near the inception of computer vision and robotic understanding~\cite{binford1982survey,chin1986model,roberts1963machine}, although constrained by the geometric models available. 
With increasing availability of larger-scale CAD model datasets~\cite{shapenet2015,sun2018pix3d}, we have seen a rejuvenation in understanding the objects of a scene by retrieving and aligning similar CAD models.
Most methods focus on aligning CAD models to RGB-D scan, point cloud geometry, or 2D-3D surface mapping though various geometric feature matching techniques~\cite{Avetisyan2019Scan2CAD,grabner2019location,izadinia2020licp,kim2013guided,li2015database,shao2012interactive}. Izadinia and Seitz~\cite{izadinia2017im2cad} and Huang et al.~\cite{huang2018holistic} propose to optimize for both scene layouts and CAD models of objects from image input, leveraging analysis-by-synthesis approaches; these methods involve costly optimization for each input image (minutes to hours).

From single image views of a object, Li et al.~\cite{li2015joint} propose a method to construct a joint embedding space between RGB images and CAD models by first constructing the space based on handcrafted shape similarity descriptors, and then optimizing for the image embeddings into the shape space. 
Our approach also optimizes for a joint embedding space between image views and CAD models in order to perform retrieval; however, we construct our space by jointly learning from both image and CAD in an end-to-end fashion without any explicit encoding of shape similarity.

\section{\OURS{}}
\label{sec:method}

\subsection{Overview}

From a single RGB image, \OURS{} detects and localizes objects by recognizing similar 3D models from a candidate set, and inferring their pose alignment to the image. 
We focus on real-world imagery and jointly learn the 2D-3D relations in an end-to-end fashion.
This produces an object-based reconstruction and understanding of the image, where each object by nature is characterized by a valid, complete 3D model with a clean, efficient geometric representation.

Specifically, from an input image, we first detect all objects in the image domain by predicting their bounding boxes, class labels, and segmentation masks. 
From these detected image regions, we then learn to construct a shared embedding space between these image regions and 3D CAD models of objects, which enables retrieving a geometrically similar model for the image observation.
We simultaneously predict the object alignment to the image as 5 dof pose optimization (z-depth translation given), yielding a 3D understanding of the objects in the image.

\paragraph{Object Detection}
For object detection, we build upon ShapeMask~\cite{kuo2019shapemask}, a state-of-the-art instance segmentation approach.
ShapeMask takes as input an RGB image and outputs detected objects characterized by their bounding boxes, class labels, and segmentation masks.
The one-stage detection approach of RetinaNet~\cite{lin2017focal} is leveraged to generate object bounding box detections, which are then refined into instance masks by a learned set of shape priors.
We modify it to leverage the learned features for our 3D shape prediction.
Each bounding box detection is used to crop features from the corresponding level of feature pyramid to produce a feature descriptor $F_i$ for the instance mask prediction $M_i$ (e.g. 32x32) of object $i$; we then take the product $M_i \circ F_i$ as the representative feature map for object $i$ as seen in the image. This is then used to inform the following CAD model retrieval and pose alignment.

\subsection{Joint Embedding Space for Image-CAD Retrieval}

The core of our approach lies in learning to seamlessly map between image views of an object and 3D CAD models, giving an association between image and 3D geometry.
The CAD models represent an explicit prior on object geometry, providing an inherently clean, complete, and compact 3D representation of an object.
We learn this mapping between image-CAD by constructing a shared embedding space -- importantly, as we show in Section~\ref{sec:Experiments}, our approach to jointly learn this embedding space constructs a space that is robust to new, unseen CAD models at test time.

Constructing a joint embedding space between image regions and 3D object geometry requires mapping across two very different domains, where in contrast to a geometric CAD model, an image is view-dependent and composed of the interaction of scene geometry with lighting and material.
To facilitate the construction of this shared space between, we thus represent each object similar to a light field descriptor~\cite{chen2003visual}, rendering a set of $k$ views ${I_0^i,...,I_k^i}$ for an object $O_i$.
For all our experiments, we use $k=16$; the set of canonical views for each object is determined by K-medoid clustering of the views seen of the object category in the training set. In addition, we augment the pool of anchor-positive pairs by using slightly jittered groundtruth views of the objects.

The embedding space is then established between the image region features $M_i\circ F_i$ from the  detection, and the 3D object renderings ${I_0^j,...,I_k^j}$.
Representative features for the image region descriptions and object renderings are extracted by a series of convolutional layers applied to each input.
The convolutional networks to extract these features are structured symmetrically, although we do not share weights due to the different input domains (See Sec. \ref{sec:details} for more details).
We refer to the resulting extracted feature descriptors as $f^{im}$ and $f^{obj}$ for the image regions and object views, respectively. The $f^{im}$ come from the regions of interest (ROI) shared with the 2D detection and segmentation branch. More specifically, the encoder backbone is a ResNet feature pyramid network and the decoder is a stack of 3x3 convolution layers on the ROI features.

We guide the construction of the embedding space with a noise contrastive estimation loss \cite{oord2018representation} for $f^{im}$ describing a detected image region
\begin{equation}
L_c = -\sum_{p \in P_{h}}\log{\frac{D(f^{im}, f^{obj}_p)}{D(f^{im}, f^{obj}_p) + C \sum_{n \in N_{h}}{D(f^{im}, f^{obj}_n)}}}
\end{equation}
where $f^{obj}_p$ represents the feature descriptor of a corresponding 3D object to the image region, $f^{obj}_n$ the feature descriptor of a non-corresponding object, $C$ a weighting parameter, and $D$ the cosine distance function:
\begin{equation}
D(x, y) \coloneqq \frac{1}{\tau}(\frac{x}{||x||})^T(\frac{y}{||y||})
\end{equation}
where $\tau$ is the temperature. $P_{h}$ and $N_{h}$ denote the set of hard positive and negative examples for the image region. Details of hard-example mining are provided in the next section. The positively corresponding objects are determined by the CAD annotations to the images, and negatively corresponding objects are the non-corresponding CAD renderings in the training batch. 

Since the object detection already provides class of the object, the negatives are only sampled from the shapes under the same class; that is, our embedding spaces are constructed for each class category although the weights are shared among them.

Empirically, we find it important to place more sampling weights on the rare classes because the number of valid pairs scale \textit{quadratically} with the number of same-class examples in the batch. We apply the inverse square root method as in \cite{gupta2019lvis} to enhance the rare class examples with a threshold $t=0.1$, which leads to improved performance on rare classes without compromising dominant classes. 

\vspace{\baselineskip}
\textbf{Hard example mining.}
Hard example mining is known to be crucial for embedding learning, as most examples are easy and do not contain much information to improve the model. We employ both hard positive and hard negative mining in \OURS{}  as follows. 
For each image region (anchor), we sample top-$P_h$ positive object views and top-$N_h$ negative object views by their distances to anchor. Similar to \cite{kuo2019shapemask}, we sample $Q$ objects for each image during the training ($Q = 8$). Since the number of objects in a batch scales linearly with $Q$, we set $P_h = 4Q = 32$ and $N_h = 16Q = 128$. Summation over hard examples allows the loss to focus on difficult cases and perform better.

\vspace{\baselineskip}
\textbf{Shape Retrieval.}
Once this embedding space is constructed, we can then perform shape retrieval at test time to provide a 3D understanding of the objects in an image.
An input image at test time is processed by the 2D detection to provide a bounding box, segmentation mask, and feature descriptor for each detected object.
We then use a nearest neighbor retrieval into the embedding space with $N_k=1$ based on cosine distance to find the most similar CAD model for each detected object. We have tried larger $N_k$ values and majority vote schemes but did not see any performance gain.

\subsection{Pose Prediction}
\label{subsec:pose_prediction}

We additionally aim to predict the pose of the retrieved 3D object such that it aligns best to the input image.
We thus propose a pose prediction branch which outputs the rotation and translation of the object. Starting with the $M_i\circ F_i$ feature map for a detected object, the object translation is directly regressed with a Huber loss~\cite{huber1992robust} as follows:
\begin{equation}
L_\delta (x) = \begin{cases}
 \frac{1}{2}{x^2}                   & \text{for } |x| \le \delta, \\
 \delta (|x| - \frac{1}{2}\delta), & \text{otherwise.}
\end{cases}
\end{equation}
The object rotation is simultaneously predicted; the rotation is first classified to a set of $K$ discretized rotation bins using cross entropy loss; this coarse estimate is then refined through a regression step using a Huber loss. This coarse-to-fine approach helps to navigate the non-euclidean rotation space, and enables continuous rotation predictions.

For rotation prediction, we represent the rotation as a quaternion, and compute the set of rotation bins by K-medoid clustering based on train object rotations. To further refine this coarse prediction, we then predict a refined rotation by estimating the delta from the classified bin using a Huber loss. The delta is represented as a $R^4$ quaternion. We initialize the bias of the last layer with $(0.95, 0, 0, 0)$ such that the quaternion is close to identity transform at the beginning. Note that during training, we only train the refinement for classified rotations within $\theta$ to avoid regressing to dissimilar targets. 

To obtain full prediction in the camera space, we need to predict the translation of the object in addition to the shape and rotation. A naive approach is to use the bounding box center as the object center in 2D and cast a ray through the center to intersects with the given groundtruth z-plane. Unfortunately, the bounding boxes tend to be unstable against the rotation and their centers can end up far from the actual object center.

We thus regress the object center as a bounding regression problem. More specifically, for each ROI, we task the network with predicting ($\delta_x$, $\delta_y$), where the $\delta$s are the shift between bounding box center and actual object center as a ratio of object width and height. 
At train time, we optimize ($\delta_x$, $\delta_y$) with the aforementioned Huber loss. At test time, we apply ($\delta_x$, $\delta_y$) to the box center to obtain the object 3D translations (assuming groundtruth depth is given \cite{meshrcnn}).

\subsection{Implementation details}
\label{sec:details}
We use ShapeMask~\cite{kuo2019shapemask} as the instance segmentation backbone. 
The model backbone is initialized from COCO-pretrained checkpoint and uses ResNet-50 architecture so as to be comparable to Mesh R-CNN in our experiments. The shape rendering branch uses a lightweight ResNet-18 backbone initialized randomly.

We freeze the weights of the backbone ResNet-50 layers after initialization and optimize both branches jointly for 48K iterations until convergence  (about 1000 epochs for Pix3D), which takes approximately 13 hours. 
The learning rate is set to be 0.08 and decays by 0.1 at 32K and 40K iterations. 
The losses for the retrieval and pose estimation are weighted with 0.5, 0.25, and 5.0 for the embedding loss, pose classification loss, and pose regression loss. 
We use $C = 1.5$ and $\tau = 0.15$ in our contrastive loss, and Huber loss margin of $\delta = 0.15$ for the pose and center regression. For pose prediction, we set $K=16$ bins, and $\theta = \pi/6$.

For each example, we randomly sample 3 out of $k=16$ canonical view renderings and one jittered groundtruth view rendering to add to the contrastive learning pool. Similar to ShapeMask, we apply ROI jittering to the image region for training the segmentation, embedding, and pose estimation branches. The noise is set to 0.025 following ShapeMask. We also apply data augmentation by horizontal image flips with 50\% probability. For such image flips, the pose labels were also adjusted accordingly.
\section{Experiments}
\label{sec:Experiments}

We evaluate our approach on the Pix3D dataset~\cite{sun2018pix3d}, which comprises $10,069$ images annotated with corresponding 3D models of the objects in the images.
We aim to jointly detect and predict the 3D shapes for the objects in the images.
We evaluate on the train/test split used by Mesh R-CNN~\cite{meshrcnn} for the same task. 
Additionally, we propose the first single-image 3D object reconstruction baseline on the ScanNet dataset~\cite{dai2017scannet}, which tends to contain more cluttered, in-the-wild views of objects.

\paragraph{Evaluation metric.}
We adopt the popular metrics from 2D object recognition, and similar to Mesh R-CNN~\cite{meshrcnn}, evaluate AP\textsuperscript{box} and AP\textsuperscript{mask} on the 2D detections, and AP\textsuperscript{mesh} on the 3D shape predictions for the objects.
Similar to Mesh R-CNN, we evaluate AP\textsuperscript{mesh} using the precision-recall for F1\textsuperscript{0.3}.
However, note that while Mesh R-CNN evaluate these metrics at IoU 0.5 (AP50), we adopt the standard COCO object detection protocol of AP50-AP95 (denoted as AP), averaging over 10 IoU thresholds of $0.5:0.05:0.95$ \cite{lin2014microsoft}.
This enables characterization of high-accuracy shape reconstructions captured at more strict IoU thresholds, demonstrating a more comprehensive description of the accuracy of the shape predictions.
In addition to AP, we also report individual AP\textsuperscript{mesh} scores for IoU thresholds of $0.5$ and $0.75$ following Mask R-CNN \cite{he2017mask}. For better reproducibility, we report every metric as an average of 5 independent runs throughout this paper.

\begin{table*}
  \centering
  \tablestyle{1.2pt}{1.1}
  \caption{Performance on Pix3D~\cite{sun2018pix3d} $\mathcal{S}_1$. 
            We report mean AP\textsuperscript{mesh} as well as per category AP\textsuperscript{mesh}. 
            AP is averaged from AP50-AP95 following the COCO detection protocol. 
            All AP performances are in \%. 
            We outperform the state-of-the-art approach on all AP metrics. 
            This improvement mostly derives from maintaining more robust performance in the high AP regime above AP50. Additionally, we observe that \OURS{} performs well on furniture categories and not so well on tools and miscellaneous objects which exhibit highly irregular shapes.
  }
  \vspace{2mm}
  \begin{tabular}{l|ccc|ccccccccc} 
     Pix3D $\mathcal{S}_1$ & \cellcolor{blue!10}AP & AP50 & AP75 & \emph{chair} & \emph{sofa} & \emph{table}   & \emph{bed}  & \emph{desk} & \emph{bkcs} &  \emph{wrdrb} &\emph{tool} & \emph{misc} \\ 
     \shline                                       
      Mesh R-CNN \cite{meshrcnn} & \cellcolor{blue!10}17.2 & 51.2 & 7.4 & 17.6 & 30.0 & 11.0 & 20.0 & 21.0 & 10.1 & 14.3 & \textbf{6.5} & \textbf{24.5} \\
    \OURS{} & \cellcolor{blue!10}\textbf{33.2} & \textbf{54.9} & \textbf{30.8} & \textbf{19.6} & \textbf{55.8} & \textbf{29.2} & \textbf{39.4} & \textbf{31.6} & \textbf{42.4} & \textbf{60.3} & 4.2 & 15.9 \\
      \hline
    \end{tabular}
   \label{table:pix3d-main}
   \vspace{-4mm}
\end{table*}
\begin{figure}
    \centering
    \includegraphics[width=\textwidth]{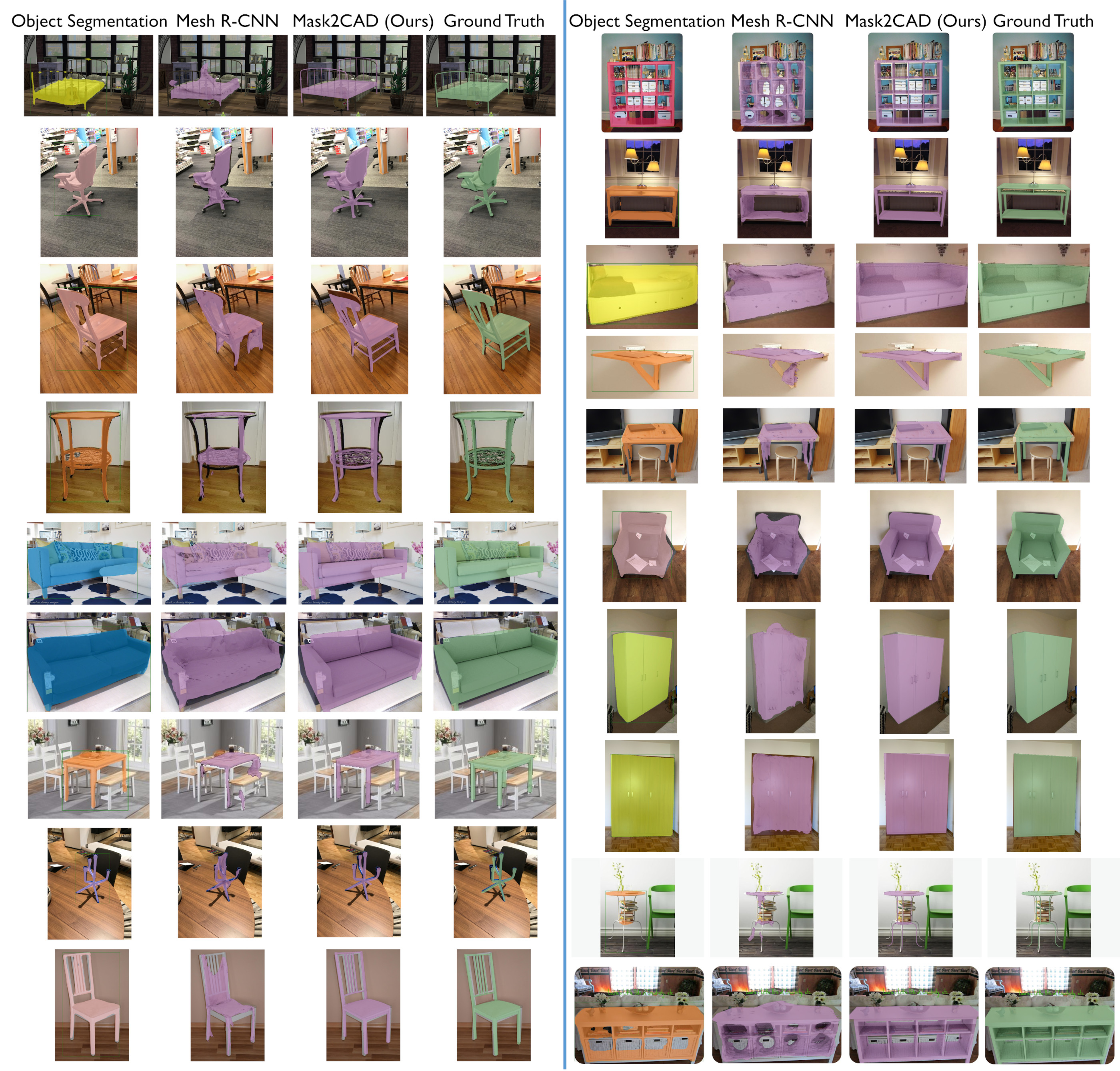}
    \caption{\OURS{} predictions on Pix3D~\cite{sun2018pix3d}. 
    The detected object is highlighted on the lefthand side of each column, with shape predictions denoted in purple, and ground truth in green.
    In contrast to Mesh R-CNN~\cite{meshrcnn}, our approach can achieve more accurate shape predictions with geometry in a clean, efficient representation.
    }
    \label{fig:pix3d}
\end{figure}

\vspace{\baselineskip}
\noindent \textbf{Comparison to state of the art.}
We compare our \OURS{} approach for 3D object understanding from RGB images by joint segmentation and retrieval to Mesh R-CNN~\cite{meshrcnn}, who first propose this task on Pix3D~\cite{sun2018pix3d}. 
Table~\ref{table:pix3d-main} shows our shape prediction results in comparison to Mesh-RCNN on their $\mathcal{S}_1$ split of the Pix3D dataset.
We evaluate AP\textsuperscript{mesh}, averaged over all class categories, as well as per-category.
In contrast to Mesh R-CNN, whose results show effective coarse predictions but suffer significantly at AP75, our shape and pose predictions maintain high-accuracy reconstructions.
We show qualitative comparisons in Figure~\ref{fig:pix3d}.

\begin{table*}
  \centering
  \tablestyle{2.5pt}{1.1}
  \caption{
  Performance on Pix3D~\cite{sun2018pix3d} with ground truth object bounding boxes given. We report Chamfer distance, Normal consistency and F1 scores. 
  Note that for these experiments, the Mesh R-CNN-based approaches are additionally provided the ground truth scale in the depth dimension of the object.
  }
  \begin{tabular}{l|ccccccccc||ccccc} 
     Pix3D $\mathcal{S}_1$ gt & Chamfer $\downarrow$ & Normal $\uparrow$ & F1$^{0.1}$ $\uparrow$ & F1$^{0.3}$ $\uparrow$& F1$^{0.5}$ $\uparrow$\\ 
     \shline
      Mask R-CNN + Pixel2Mesh \cite{meshrcnn} & 1.30 & 0.70 & 16.4 & 51.0 & 68.4 \\
      Mesh R-CNN (Voxel-Only) \cite{meshrcnn} & 1.28 & 0.57 & 9.9 & 37.3 & 56.1 \\
      Mesh R-CNN (Sphere-Init) \cite{meshrcnn} & 1.30 & 0.69 & 16.8 & 51.4 & 68.8 \\
      Mesh R-CNN \cite{meshrcnn} & 1.11 & 0.71 & 18.7 & 56.4 & 73.5 \\
      \hline
      \OURS{} (Ours) & \textbf{0.99} &  \textbf{0.74} & \textbf{25.6} & \textbf{66.4} & \textbf{79.3} \\
   \end{tabular}
  \label{table:pix3d-gt}
\end{table*}

\begin{comment}
  \resizebox{\textwidth}{!}{  
  \begin{tabular}{l|ccc|ccccccccc||ccccc} 
     Pix3D $\mathcal{S}_1$ gt & AP50$^{\text{box}}$  & AP50$^{\text{mask}}$ & AP50$^{\text{mesh}}$ & Chamfer & Normal & F1$^{0.1}$ & F1$^{0.3}$ & F1$^{0.5}$ \\ 
     \shline
      Mask R-CNN + Pixel2Mesh \cite{meshrcnn} & 100.0 & 92.0 & 35.1 & 1.30 & 0.70 & 16.4 & 51.0 & 68.4 \\
      Mesh R-CNN (Voxel-Only) \cite{meshrcnn} & 100.0 & 90.7 & 6.7 & 1.28 & 0.57 & 9.9 & 37.3 & 56.1 \\
      Mesh R-CNN (Sphere-Init) \cite{meshrcnn} & 100.0 & 92.4 & 33.4 & 1.30 & 0.69 & 16.8 & 51.4 & 68.8 \\
      Mesh R-CNN \cite{meshrcnn} & 100.0 & 92.1 & \textbf{49.1} & \textbf{1.11} & \textbf{0.71} & 18.7 & 56.4 & \textbf{73.5} \\
      \hline
      \OURS{} (Ours) & 100.0 & 93.9 & 42.9 & 1.26 & \textbf{0.71} & \textbf{21.5} & \textbf{59.5} & 73.2 \\
   \end{tabular}
   }
\end{comment}

Additionally, we compare to several state-of-the-art single object reconstruction approaches on Pix3D $\mathcal{S}_1$ in Table~\ref{table:pix3d-gt}; for each approach we provide ground truth 2D object detections, i.e. perfect bounding boxes.
We evaluate various characteristics of the shape reconstruction. 
We also evaluate the Chamfer distance, normal consistency, and F1 at thresholds 0.1, 0.3, 0.5, using randomly sampled points on the predicted and ground truth meshes, where meshes are scaled such that the longest edge of the ground-truth mesh’s bounding box has length 10.
Chamfer distance and normal consistency provide more global measures of shape consistency with the ground truth, but can tend towards favoring averaging, while F1 scores tend to be more robust towards outliers, and F1 at lower thresholds in particular indicates the ability to predict highly-accurate shapes.
Note the competing approaches have been provided the ground truth scale in the depth dimension at test time, while our approach directly retrieves it from the training set.
Nonetheless, our approach can provide higher-accuracy predictions as seen in the F1 scores at 0.1 and 0.3.

\begin{figure}[t]
    \centering
    \includegraphics[width=0.89\textwidth]{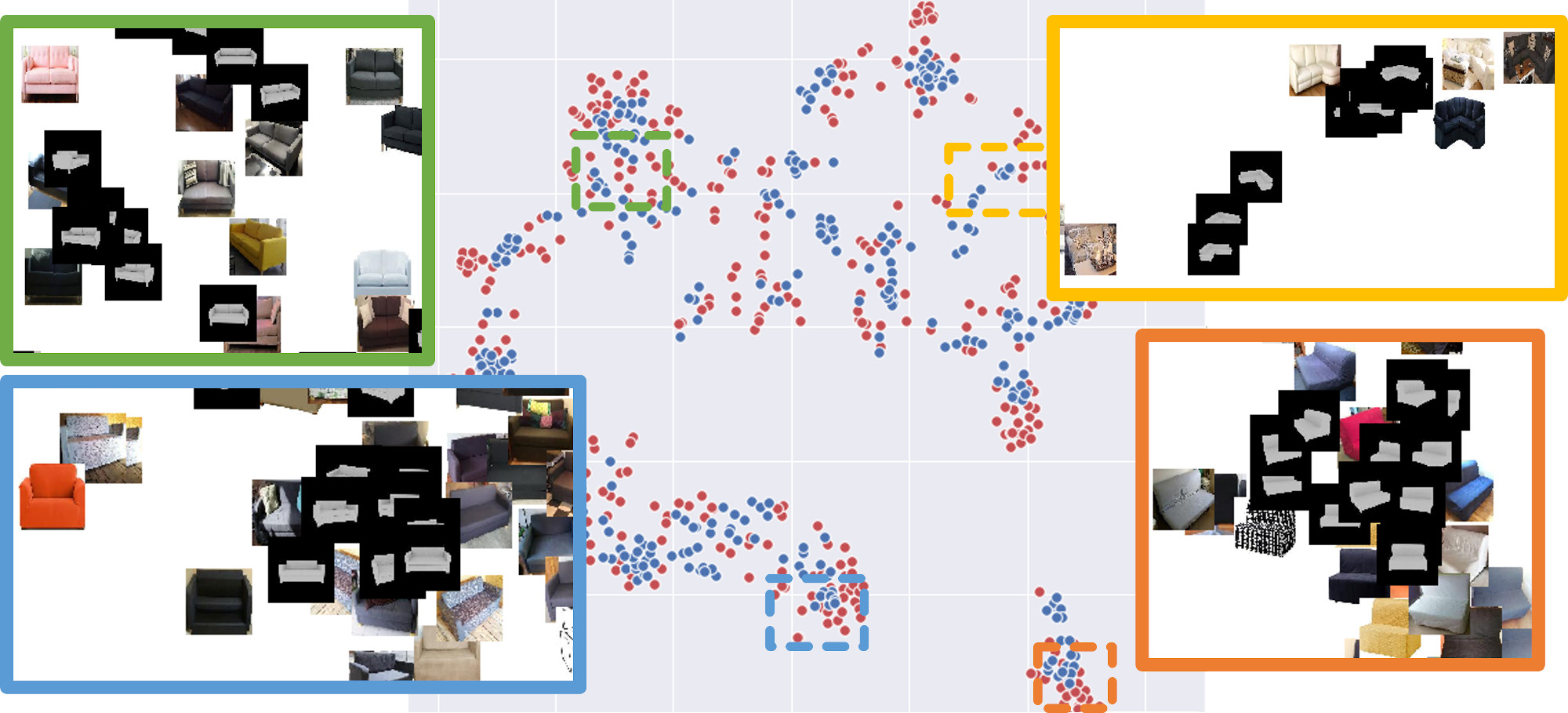} 
    \includegraphics[width=0.89\textwidth]{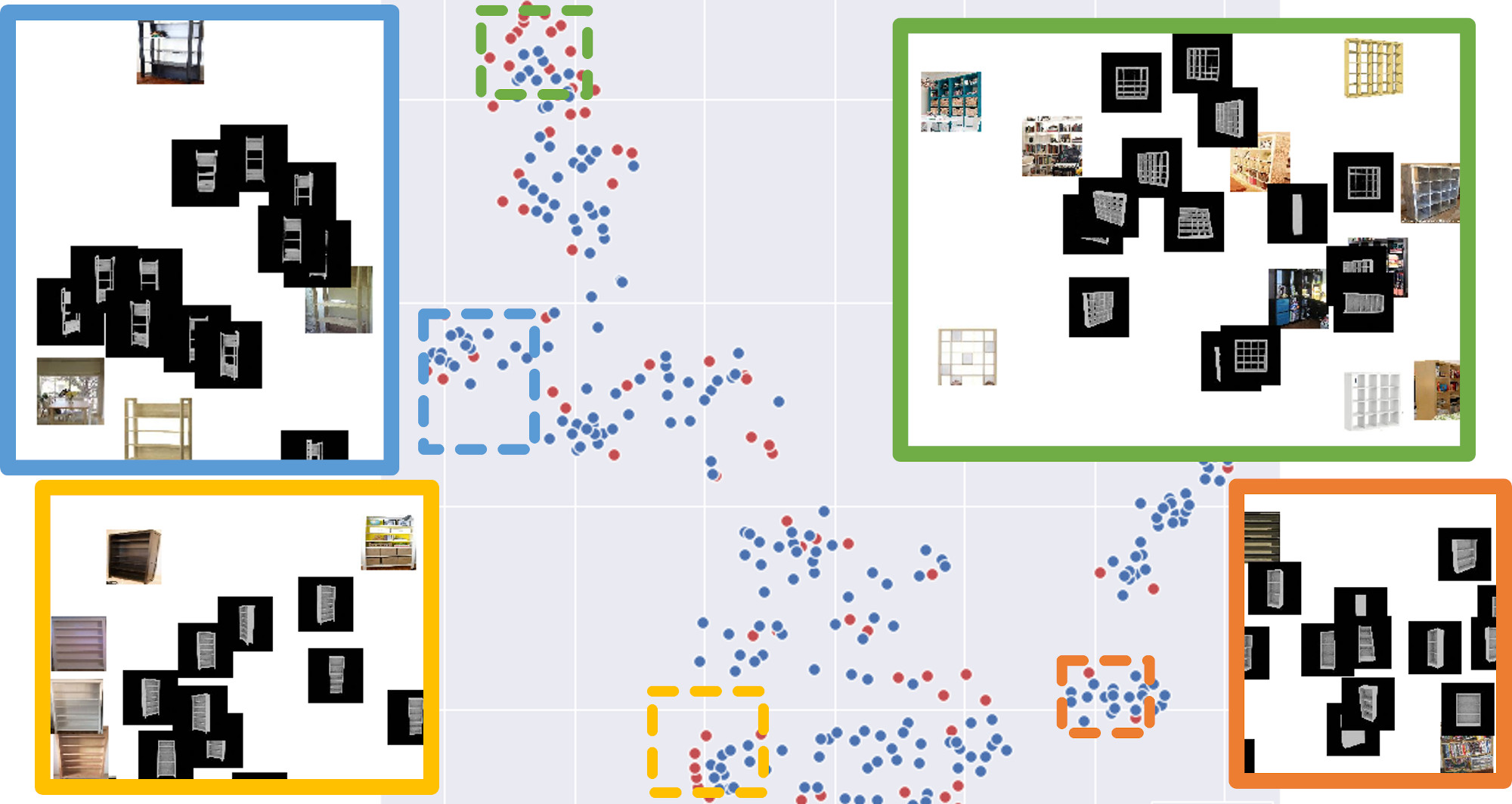} 
    \caption{t-SNE embeddings of \OURS{} for the sofa (top) and bookcase (bottom) classes. More visualizations can be found in the Supp. Materials. 
    Red points correspond to images, and blue to shapes.
    Both images and shapes mix well together in the embedding space.
    Note that despite lack of shape similarity information during training, clusters tend to form together in geometric similarity,  e.g., L-shaped sofas (yellow), single seat sofas without armrests (orange), single seat sofas with armrests (blue), double seat sofas (green). 
    This stands in contrast to the embedding space construction of ~\cite{li2015joint} which explicitly enforces shape similarity in its light field descriptors.
    }
    \vspace{-4mm}
    \label{fig:cadmask_tsne}
\end{figure}

\vspace{\baselineskip}
\noindent \textbf{Implicit learning of shape similarity.}
In Figure~\ref{fig:cadmask_tsne}, we visualize our learned embedding space by t-SNE~\cite{maaten2008visualizing}, for image regions and CAD models of the \emph{sofa} and \emph{bookcase} class categories (we refer to the supplemental material for additional visualizations of the learned embedding spaces).
We find that not only do the images and shapes mix together in this embedding space, despite that it is constructed without any knowledge of shape similarity -- only image-CAD associations --, geometrically similar shapes tend to cluster together. 

\vspace{\baselineskip}
\noindent \textbf{Can the image-shape embedding space generalize to new 3D models?}
Our joint image-CAD model embedding space constructed during train time leverages ground truth annotations of CAD models to images, which can be costly to acquire.
During inference time, however, we can still embed new 3D models into the space without training, by using our trained model to compute their feature embeddings.
Our embedding approach generalizes well in incorporating these new models.

We demonstrate this on the $\mathcal{S}_2$ split of Pix3D, training on a subset of the 3D CAD models, with test images comprising views of a disjoint set of objects than those in the training set. 
Generalization under this regime is difficult, particularly for a retrieval-based approach.
However, in Table~\ref{table:pix3d-s2-gen} we show clear improvements when using all available CAD models at test time in comparison to only the CAD models in the train set, despite not having seen any of the new objects nor their corresponding image views.

To help the model generalize better, we apply more data augmentation than the $\mathcal{S}_1$ split, including HSV-space jittering, random crop and resize of the renderings, and augmenting the box and image jittering magnitude as used in ShapeMask \cite{kuo2019shapemask}.

\begin{table*}
  \centering
  \tablestyle{1.2pt}{1.1}
  \caption{Test-time generalization on Pix3D \cite{sun2018pix3d} $\mathcal{S}_2$. The performance improves on all categories with the addition 139 of CAD models at test time without re-training. 
  }
  \vspace{-1mm}
  \resizebox{\textwidth}{!}{ 
  \begin{tabular}{l|ccc|ccccccccc} 
     Pix3D $\mathcal{S}_2$ & AP & AP50 & AP75 & \emph{chair} & \emph{sofa} & \emph{table}   & \emph{bed}  & \emph{desk} & \emph{bkcs} &  \emph{wrdrb} &\emph{tool} & \emph{misc} \\ 
     \shline                                       
      \OURS{} (Ours) & 6.5 & 17.3 & 3.8 & 3.2 & 35.4 & 1.2 & 14.0 & 0.2 & 2.2 & 1.6 & 0.6 & 0.0 \\
      \OURS{} (Ours) + CAD & \textbf{8.2} & \textbf{20.7} & \textbf{4.8} & \textbf{4.5} & \textbf{37.8} & \textbf{3.6} & \textbf{16.9} & \textbf{2.7} & \textbf{2.2} & \textbf{5.3} & \textbf{0.9} & \textbf{0.1} \\
      \hline
    \end{tabular}
    }
   \label{table:pix3d-s2-gen}
   \vspace{-4mm}
\end{table*}

\begin{figure}
    \centering
    \includegraphics[width=1.0\textwidth]{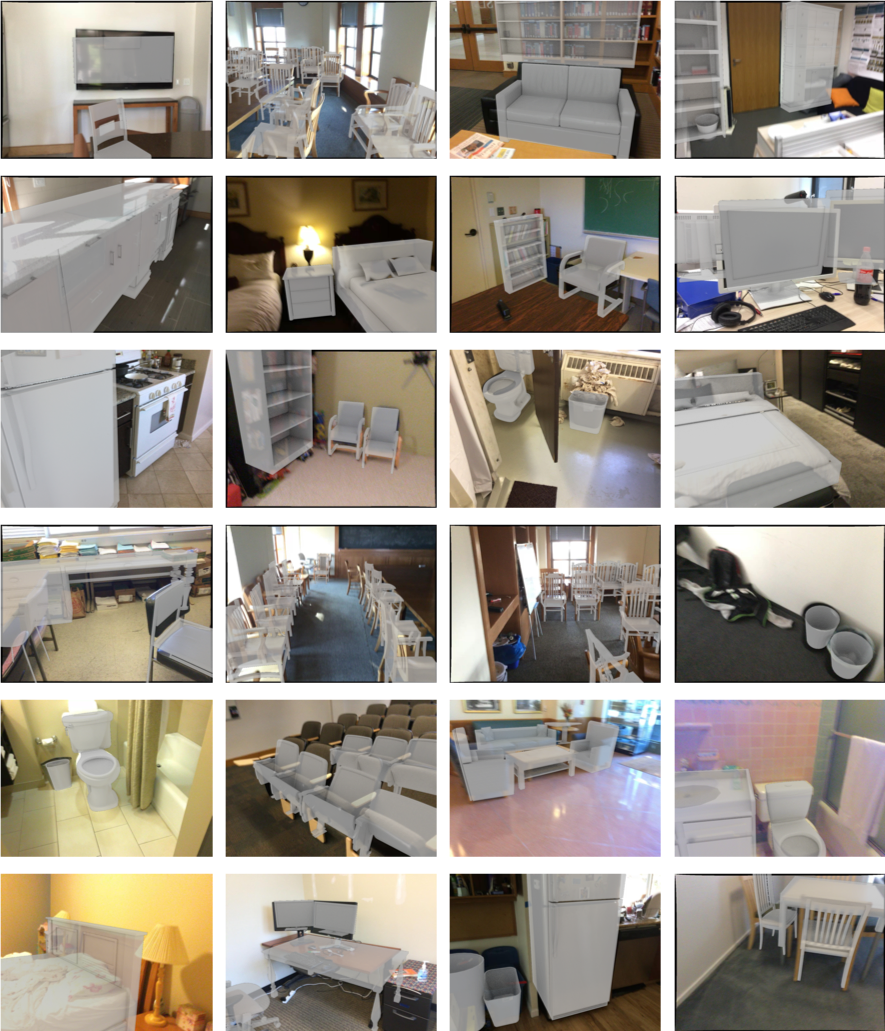}
    \caption{Example \OURS{} predictions on ScanNet~\cite{dai2017scannet} images. 
    Our approach shows encouraging results in its application to the more diverse set of image views, lighting, occlusions, and object categories of ScanNet.
    }
    \label{fig:scannet}
\end{figure}

\vspace{\baselineskip}
\noindent \textbf{Comparison with ShapeNet reconstruction methods.}
In Table \ref{table:pix3d-benchmark}, we compare the Pix3D $\mathcal{S}_1$ model on the validation set with the other methods that train on ShapeNet with real data augmentation. The evaluation protocol and implementation follows \cite{sun2018pix3d}. \OURS{} results are reported on 1165 chairs in the $\mathcal{S}_1$ test split of Mesh R-CNN \cite{meshrcnn}, as an average over 5 independent runs. Surprisingly, \OURS{} achieves significantly better shape predictions than the state-of-the-art methods (0.288 IoU, and 0.013 Chamfer Distance), showing the capability of retrieval. 
\begin{table*}
  \centering
  \tablestyle{1.2pt}{1.5}
  \caption{\OURS{} on Pix3D~\cite{sun2018pix3d} $\mathcal{S}_1$ test split in comparison with other methods that train on ShapeNet models with real data augmentation.
  }
  \begin{tabular}{l|c|c|c|c|c} 
     & \OURS{} (Ours) & FroDO \cite{li2020frodo} & Sun et al. \cite{sun2018pix3d} & MarrNet \cite{wu2017marrnet} & 3D-R2N2 \cite{choy20163d} \\ 
     \shline                                       
    IoU  & \textbf{0.613} &  0.325 & 0.282 & 0.231 & 0.136 \\
     \hline 
    Chamfer & \textbf{0.086} & 0.099 & 0.119 & 0.144 & 0.239 \\
      \hline
    \end{tabular}
   \label{table:pix3d-benchmark}
\end{table*}

\vspace{\baselineskip}
\noindent \textbf{Baseline on ScanNet Dataset.}
We additionally apply \OURS{} to real-world images from the ScanNet dataset~\cite{dai2017scannet}, which contains RGB-D video data of $1513$ indoor scenes. We use the 25K frame subset provided by the dataset for training and validation. The train/val split contains $19387$/$5436$ images respectively, and the images come from separate scenes with distinct objects. Compared to Pix3D, this dataset has an order of magnitude more shapes, as well as many more occlusions, diverse image views, lighting conditions, and importantly, metric 3D groundtruth of the scene. We believe this could be a suitable benchmark for object 3D prediction from a real single image. 

We use the CAD labels from Scan2CAD \cite{Avetisyan2019Scan2CAD} by projecting the CAD models to each image view and use the amodal box, mask, pose, and shape for training. We additionally remove the objects whose centers are out of frame from training and evaluation. We also remove the object categories that appear less than 1000 times in the training split, resulting in eight categories: bed, sofa, chair (inc. toilet), bin, cabinet (inc. fridge), display, table, and bookcase. Regarding shape similarity, we adopt F score = 0.5 as the threshold for Mesh AP computation, because the Scan2CAD annotations come from ShapeNet and do not provide exact matches to the images. As Scan2CAD provides 9-DoF annotation for each object, we apply the groundtruth z depth and (x, y, z) scale to the predicted shape before computing the shape similarity metrics. We trained \OURS{} for 72000 iterations with HSV-color, ROI, and image scale jittering using the same learning rate schedule as Pix3D. The quantitative results are reported in Table \ref{table:scannet-main}. Despite the complexity of ScanNet data, \OURS{} manages to recognize the object shapes in these images, as shown in Figure~\ref{fig:scannet}. Our CAD model retrieval and alignment shows promising results and a potential for facilitating content creation pipelines.

\begin{table*}
  \centering
  \tablestyle{1.5pt}{1.5}
  \caption{Performance on ScanNet ~\cite{dai2017scannet}. 
        We report mean AP\textsuperscript{mesh} as well as per category AP\textsuperscript{mesh} following Pix3D protocol.
  }
  \vspace{2mm}
  \begin{tabular}{l|ccc|cccccccc} 
     ScanNet 25K & \cellcolor{blue!10}AP & AP50 & AP75 & \emph{bed} & \emph{sofa} & \emph{chair} & \emph{cab}  & \emph{bin} & \emph{disp} & \emph{table} &\emph{bkcs}  \\ 
     \shline
    \OURS{}(Ours) & \cellcolor{blue!10}8.4 & 23.1 & 4.9 & 14.2 & 13.0 & 13.2 & 7.5 & 7.8 & 5.9 & 2.9 & 3.1 \\
      \hline
    \end{tabular}
   \label{table:scannet-main}
   \vspace{-4mm}
\end{table*}

\vspace{\baselineskip}
\noindent \textbf{Runtime.}
At test time \OURS{} is efficient and runs at approximately 60 milliseconds per 640 by 640 image on Pix3D, including 2D detection and segmentation as well as shape retrieval and pose estimation.

\paragraph{Limitations.}
While \OURS{} shows promising progress in attaining 3D understanding of the objects from a single image, we believe there are many avenues for further development.
For instance, our retrieval-based approach can suffer in the case of objects that differ too strongly from the existing CAD model database, and we believe that mesh-based approaches to deform and refine geometry~\cite{tang2019skeleton,wang2018pixel2mesh} have significant potential to complement our approach.
Additionally, we believe a holistic 3D scene understanding characterizing not only the objects in an environment but all elements in the scene is a promising direction for 3D perception and semantic understanding.

\section{Conclusion}

We propose \OURS{}, a novel approach for 3D perception from 2D images. Our method leverages a CAD model representation, and jointly detects objects for an input image and retrieves and aligns a similar CAD model to the detected region.
We show that our approach produces accurate shape reconstructions and is capable of generalizing to unseen 3D objects at test time.
The final output of \OURS{} is a CAD-based object understanding of the image, where each object is represented in a clean, lightweight fashion.
We believe that this makes a promising step in 3D perception from images as well as transforming real-world imagery to a synthetic representation, opening up new possibilities for digitization of real-world environments for applications such as content creation or domain transfer.

\section*{Acknowledgments}
We would like to thank Georgia Gkioxari for her advice on Mesh R-CNN and the support of the ZD.B (Zentrum Digitalisierung.Bayern) for Angela Dai.

\clearpage
\bibliographystyle{splncs04}
\bibliography{egbib}

\begin{thebibliography}{10}
\providecommand{\url}[1]{\texttt{#1}}
\providecommand{\urlprefix}{URL }
\providecommand{\doi}[1]{https://doi.org/#1}

\bibitem{Avetisyan2019Scan2CAD}
Avetisyan, A., Dahnert, M., Dai, A., Savva, M., Chang, A.X., Nie{\ss}ner, M.:
  Scan2cad: Learning cad model alignment in rgb-d scans. CVPR  (2019)

\bibitem{binford1982survey}
Binford, T.O.: Survey of model-based image analysis systems. The International
  Journal of Robotics Research  \textbf{1}(1),  18--64 (1982)

\bibitem{shapenet2015}
Chang, A.X., Funkhouser, T., Guibas, L., Hanrahan, P., Huang, Q., Li, Z.,
  Savarese, S., Savva, M., Song, S., Su, H., Xiao, J., Yi, L., Yu, F.:
  {ShapeNet: An Information-Rich 3D Model Repository}. Tech. Rep.
  arXiv:1512.03012 [cs.GR], Stanford University --- Princeton University ---
  Toyota Technological Institute at Chicago (2015)

\bibitem{chen2003visual}
Chen, D.Y., Tian, X.P., Shen, Y.T., Ouhyoung, M.: On visual similarity based 3d
  model retrieval. In: Computer graphics forum. vol.~22, pp. 223--232. Wiley
  Online Library (2003)

\bibitem{chin1986model}
Chin, R.T., Dyer, C.R.: Model-based recognition in robot vision. ACM Computing
  Surveys (CSUR)  \textbf{18}(1),  67--108 (1986)

\bibitem{choy20163d}
Choy, C.B., Xu, D., Gwak, J., Chen, K., Savarese, S.: 3d-r2n2: A unified
  approach for single and multi-view 3d object reconstruction. In: European
  conference on computer vision. pp. 628--644. Springer (2016)

\bibitem{dai2017scannet}
Dai, A., Chang, A.X., Savva, M., Halber, M., Funkhouser, T., Nie{\ss}ner, M.:
  {ScanNet}: Richly-annotated {3D} reconstructions of indoor scenes. In: Proc.
  Computer Vision and Pattern Recognition (CVPR), IEEE (2017)

\bibitem{dai2019scan2mesh}
Dai, A., Nie{\ss}ner, M.: Scan2mesh: From unstructured range scans to 3d
  meshes. In: Proceedings of the IEEE Conference on Computer Vision and Pattern
  Recognition. pp. 5574--5583 (2019)

\bibitem{duan2019centernet}
Duan, K., Bai, S., Xie, L., Qi, H., Huang, Q., Tian, Q.: Centernet: Keypoint
  triplets for object detection. In: Proceedings of the IEEE International
  Conference on Computer Vision. pp. 6569--6578 (2019)

\bibitem{fan2017point}
Fan, H., Su, H., Guibas, L.J.: A point set generation network for 3d object
  reconstruction from a single image. In: Proceedings of the IEEE conference on
  computer vision and pattern recognition. pp. 605--613 (2017)

\bibitem{girshick2015fast}
Girshick, R.: Fast r-cnn. In: Proceedings of the IEEE international conference
  on computer vision. pp. 1440--1448 (2015)

\bibitem{meshrcnn}
Gkioxari, G., Malik, J., Johnson, J.: Mesh r-cnn. arXiv preprint
  arXiv:1906.02739  (2019)

\bibitem{grabner2019location}
Grabner, A., Roth, P.M., Lepetit, V.: Location field descriptors: Single image
  3d model retrieval in the wild. In: 2019 International Conference on 3D
  Vision (3DV). pp. 583--593. IEEE (2019)

\bibitem{groueix2018papier}
Groueix, T., Fisher, M., Kim, V.G., Russell, B.C., Aubry, M.: A
  papier-m{\^a}ch{\'e} approach to learning 3d surface generation. In:
  Proceedings of the IEEE conference on computer vision and pattern
  recognition. pp. 216--224 (2018)

\bibitem{gupta2019lvis}
Gupta, A., Dollar, P., Girshick, R.: Lvis: A dataset for large vocabulary
  instance segmentation. In: Proceedings of the IEEE Conference on Computer
  Vision and Pattern Recognition. pp. 5356--5364 (2019)

\bibitem{he2017mask}
He, K., Gkioxari, G., Doll{\'a}r, P., Girshick, R.: Mask r-cnn. In: Computer
  Vision (ICCV), 2017 IEEE International Conference on. pp. 2980--2988. IEEE
  (2017)

\bibitem{huang2018holistic}
Huang, S., Qi, S., Zhu, Y., Xiao, Y., Xu, Y., Zhu, S.C.: Holistic {3D} scene
  parsing and reconstruction from a single {RGB} image. In: European Conference
  on Computer Vision. pp. 194--211. Springer (2018)

\bibitem{huber1992robust}
Huber, P.J.: Robust estimation of a location parameter. In: Breakthroughs in
  statistics, pp. 492--518. Springer (1992)

\bibitem{izadinia2020licp}
Izadinia, H., Seitz, S.M.: Scene recomposition by learning-based icp. In: CVPR
  (2020)

\bibitem{izadinia2017im2cad}
Izadinia, H., Shan, Q., Seitz, S.M.: Im2cad. In: Proceedings of the IEEE
  Conference on Computer Vision and Pattern Recognition. pp. 5134--5143 (2017)

\bibitem{kim2013guided}
Kim, Y.M., Mitra, N.J., Huang, Q., Guibas, L.: Guided real-time scanning of
  indoor objects. In: Computer Graphics Forum. vol.~32, pp. 177--186. Wiley
  Online Library (2013)

\bibitem{kuo2019shapemask}
Kuo, W., Angelova, A., Malik, J., Lin, T.Y.: Shapemask: Learning to segment
  novel objects by refining shape priors. In: Proceedings of the IEEE
  International Conference on Computer Vision. pp. 9207--9216 (2019)

\bibitem{li2020frodo}
Li, K., R{\"u}nz, M., Tang, M., Ma, L., Kong, C., Schmidt, T., Reid, I.,
  Agapito, L., Straub, J., Lovegrove, S., et~al.: Frodo: From detections to 3d
  objects. arXiv preprint arXiv:2005.05125  (2020)

\bibitem{li2015database}
Li, Y., Dai, A., Guibas, L., Nie{\ss}ner, M.: Database-assisted object
  retrieval for real-time 3d reconstruction. In: Computer Graphics Forum.
  vol.~34, pp. 435--446. Wiley Online Library (2015)

\bibitem{li2015joint}
Li, Y., Su, H., Qi, C.R., Fish, N., Cohen-Or, D., Guibas, L.J.: Joint
  embeddings of shapes and images via cnn image purification. ACM transactions
  on graphics (TOG)  \textbf{34}(6),  1--12 (2015)

\bibitem{lin2017focal}
Lin, T.Y., Goyal, P., Girshick, R., He, K., Doll{\'a}r, P.: Focal loss for
  dense object detection. In: Proceedings of the IEEE international conference
  on computer vision. pp. 2980--2988 (2017)

\bibitem{lin2014microsoft}
Lin, T.Y., Maire, M., Belongie, S., Hays, J., Perona, P., Ramanan, D.,
  Doll{\'a}r, P., Zitnick, C.L.: Microsoft coco: Common objects in context. In:
  European conference on computer vision. pp. 740--755. Springer (2014)

\bibitem{liu2016ssd}
Liu, W., Anguelov, D., Erhan, D., Szegedy, C., Reed, S., Fu, C.Y., Berg, A.C.:
  Ssd: Single shot multibox detector. In: European conference on computer
  vision. pp. 21--37. Springer (2016)

\bibitem{maaten2008visualizing}
Maaten, L.v.d., Hinton, G.: Visualizing data using t-sne. Journal of machine
  learning research  \textbf{9}(Nov),  2579--2605 (2008)

\bibitem{mescheder2019occupancy}
Mescheder, L., Oechsle, M., Niemeyer, M., Nowozin, S., Geiger, A.: Occupancy
  networks: Learning 3d reconstruction in function space. In: Proceedings of
  the IEEE Conference on Computer Vision and Pattern Recognition. pp.
  4460--4470 (2019)

\bibitem{oord2018representation}
Oord, A.v.d., Li, Y., Vinyals, O.: Representation learning with contrastive
  predictive coding. arXiv preprint arXiv:1807.03748  (2018)

\bibitem{park2019deepsdf}
Park, J.J., Florence, P., Straub, J., Newcombe, R., Lovegrove, S.: Deepsdf:
  Learning continuous signed distance functions for shape representation. In:
  Proceedings of the IEEE Conference on Computer Vision and Pattern
  Recognition. pp. 165--174 (2019)

\bibitem{redmon2016you}
Redmon, J., Divvala, S., Girshick, R., Farhadi, A.: You only look once:
  Unified, real-time object detection. In: Proceedings of the IEEE conference
  on computer vision and pattern recognition. pp. 779--788 (2016)

\bibitem{ren2015faster}
Ren, S., He, K., Girshick, R., Sun, J.: Faster r-cnn: Towards real-time object
  detection with region proposal networks. In: Advances in neural information
  processing systems. pp. 91--99 (2015)

\bibitem{riegler2017octnet}
Riegler, G., Osman~Ulusoy, A., Geiger, A.: Octnet: Learning deep 3d
  representations at high resolutions. In: Proceedings of the IEEE Conference
  on Computer Vision and Pattern Recognition. pp. 3577--3586 (2017)

\bibitem{roberts1963machine}
Roberts, L.G.: Machine perception of three-dimensional solids. Ph.D. thesis,
  Massachusetts Institute of Technology (1963)

\bibitem{shao2012interactive}
Shao, T., Xu, W., Zhou, K., Wang, J., Li, D., Guo, B.: An interactive approach
  to semantic modeling of indoor scenes with an rgbd camera. ACM Transactions
  on Graphics (TOG)  \textbf{31}(6),  1--11 (2012)

\bibitem{sun2018pix3d}
Sun, X., Wu, J., Zhang, X., Zhang, Z., Zhang, C., Xue, T., Tenenbaum, J.B.,
  Freeman, W.T.: {Pix3D}: Dataset and methods for single-image {3D} shape
  modeling. In: Proceedings of the IEEE Conference on Computer Vision and
  Pattern Recognition. pp. 2974--2983 (2018)

\bibitem{tang2019skeleton}
Tang, J., Han, X., Pan, J., Jia, K., Tong, X.: A skeleton-bridged deep learning
  approach for generating meshes of complex topologies from single rgb images.
  In: Proceedings of the IEEE Conference on Computer Vision and Pattern
  Recognition. pp. 4541--4550 (2019)

\bibitem{tatarchenko2017octree}
Tatarchenko, M., Dosovitskiy, A., Brox, T.: Octree generating networks:
  Efficient convolutional architectures for high-resolution 3d outputs. In:
  Proceedings of the IEEE International Conference on Computer Vision. pp.
  2088--2096 (2017)

\bibitem{wang2018pixel2mesh}
Wang, N., Zhang, Y., Li, Z., Fu, Y., Liu, W., Jiang, Y.G.: Pixel2mesh:
  Generating 3d mesh models from single rgb images. In: Proceedings of the
  European Conference on Computer Vision (ECCV). pp. 52--67 (2018)

\bibitem{wu2017marrnet}
Wu, J., Wang, Y., Xue, T., Sun, X., Freeman, B., Tenenbaum, J.: Marrnet: 3d
  shape reconstruction via 2.5 d sketches. In: Advances in neural information
  processing systems. pp. 540--550 (2017)

\bibitem{wu2016learning}
Wu, J., Zhang, C., Xue, T., Freeman, B., Tenenbaum, J.: Learning a
  probabilistic latent space of object shapes via 3d generative-adversarial
  modeling. In: Advances in neural information processing systems. pp. 82--90
  (2016)

\bibitem{yang2019pointflow}
Yang, G., Huang, X., Hao, Z., Liu, M.Y., Belongie, S., Hariharan, B.:
  Pointflow: 3d point cloud generation with continuous normalizing flows. In:
  Proceedings of the IEEE International Conference on Computer Vision. pp.
  4541--4550 (2019)

\end{thebibliography}

\newpage
\section*{Appendix A: Additional Results on Pix3D}
\begin{figure}
    \centering
    \includegraphics[width=\textwidth]{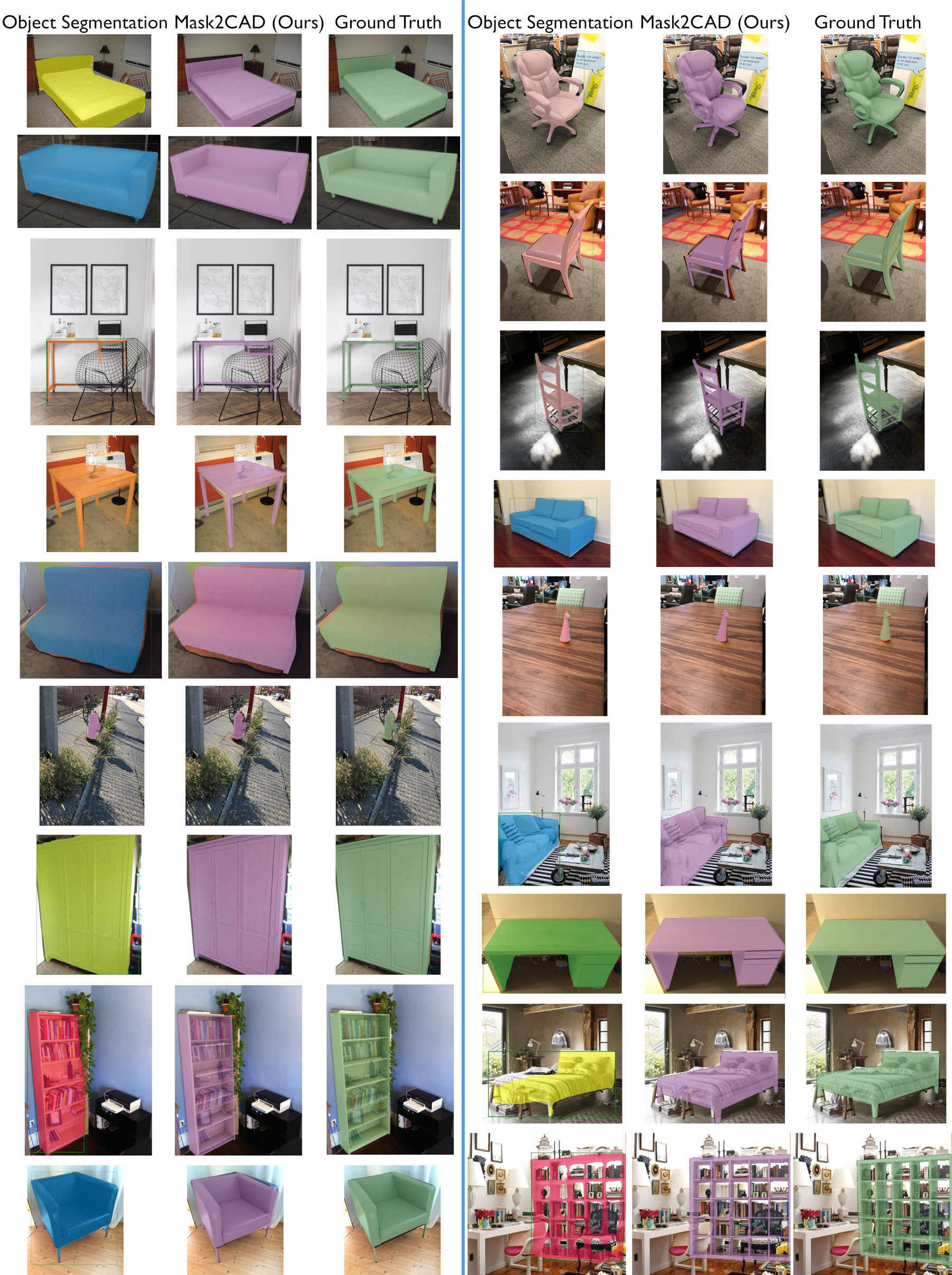} \\
    \caption{Additional qualitative results of \OURS{} on Pix3D~\cite{sun2018pix3d}.}
    \label{fig:pix3d_ours}
\end{figure}

In Figure~\ref{fig:pix3d_ours}, we show more qualitative results of our approach on Pix3D~\cite{sun2018pix3d}. Furthermore, we conduct ablation studies to shed light on the roles of each component in the system. Our analysis shows that shape, pose, and translation are all important for estimating the viewer-centered geometry, with shape retrieval having the most room for improvement, and box detection having the least. The analysis was done by replacing each predicted component with its ground truth counterpart. In terms of Mesh AP, groundtruth shapes help by +14.6, rotation by +10.2, and translation by +7.5. Surprisingly, groundtruth 2D boxes offer no improvement because the detections on Pix3D are very good (~90 Box AP, similar to Mesh R-CNN) and the small advantage is offset by the distribution shift between train (jittered boxes) and test (perfect boxes) time. This agrees with what Mesh R-CNN reports, i.e. they also observed a loss when using ground truth boxes (6 point loss on Mesh AP50).

\section*{Appendix B: Network Architecture Details}

The \OURS{} image-stream network architecture comprises 2D detection as bounding box, class label, and instance mask prediction, as well as our 3D shape retrieval and pose estimation.
For the 2D detection, our architecture borrows from that of ShapeMask~\cite{kuo2019shapemask}.
For the 3D inference with shape embedding, pose classification, and pose regression, and object center prediction, these branches all use the same architecture as the coarse mask prediction branch of \cite{kuo2019shapemask} (with the exception of the output layers). 
The inputs of these branches are the features from the region of interest (ROI) of detection backbone feature pyramid network. 
We detail each branch in Table \ref{table:arch-shape-embed}, \ref{table:arch-pose-class}, and \ref{table:arch-center-reg}.

\begin{table}
  \centering
  \scalebox{0.75}{
  \begin{tabular}{|c|c|l|c|}
    \hline
    \textbf{Index} & \textbf{Inputs} & \textbf{Operation} & \textbf{Output shape} \\
    \hline
    (1) & Input & Region of Interest (ROI) features & $32\times32\times256$ \\
    (2) & (1) & $3\times$ (Conv($256\rightarrow 256, 3\times3$) + BatchNorm + ReLU) & $32\times32\times256$ \\
    (3) & (2) & (Conv($256\rightarrow 256, 3\times3$) + BatchNorm + ReLU) & $32\times32\times128$ \\
    (4) & (3) & AveragePool(axes=(0, 1)) & $128$ \\
    \hline
  \end{tabular}}
  \vspace{2mm}
  \caption{Network architecture of the shape embedding branch. The last convolution layer downsamples the number of channels from 256 to 128.
  }
  \label{table:arch-shape-embed}
  \vspace{-2mm}
\end{table}

\begin{table}
  \centering
  \scalebox{0.75}{
  \begin{tabular}{|c|c|l|c|}
    \hline
    \textbf{Index} & \textbf{Inputs} & \textbf{Operation} & \textbf{Output shape} \\
    \hline
    (1) & Input & Region of Interest (ROI) features & $32\times32\times256$ \\
    (2) & (1) & $4\times$ (Conv($256\rightarrow 256, 3\times3$) + BatchNorm + ReLU) & $32\times32\times256$ \\
    (3) & (2) & AveragePool(axes=(0, 1)) & $256$ \\
    (4) & (3) & Linear($256\rightarrow N_{pose} \times N_{class}$) & $160$ \\
    \hline
  \end{tabular}}
  \vspace{2mm}
  \caption{Network architecture of the pose prediction branch. 
  For pose classification, the output is $N_{pose}=16$ for each class $N_{class}=10$. 
  For the following pose regression after this classification, the architecture is identical except for using $N_{pose}=4$ for predicting the regression quaternion instead of the $16$ medoid bins.
  }
  \label{table:arch-pose-class}
\end{table}

\begin{table}
  \centering
  \scalebox{0.75}{
  \begin{tabular}{|c|c|l|c|}
    \hline
    \textbf{Index} & \textbf{Inputs} & \textbf{Operation} & \textbf{Output shape} \\
    \hline
    (1) & Input & Region of Interest (ROI) features & $32\times32\times256$ \\
    (2) & (1) & $4\times$ (Conv($256\rightarrow 256, 3\times3$) + BatchNorm + ReLU) & $32\times32\times256$ \\
    (3) & (2) & AveragePool(axes=(0, 1)) & $256$ \\
    (4) & (3) & Linear($256\rightarrow N_{center} \times N_{class}$) & $20$ \\
    \hline
  \end{tabular}}
  \vspace{2mm}
  \caption{Network architecture of the object center regression branch. The output is $N_{center}=2$ for each class $N_{class}=10$, where $N_{center}$ equals 2 for ($\delta_x$, $\delta_y$).
  }
  \label{table:arch-center-reg}
  \vspace{-2mm}
\end{table}

\section*{Appendix C: t-SNE visualizations for image-CAD embeddings}
\begin{figure}
    \centering
    \includegraphics[width=\textwidth]{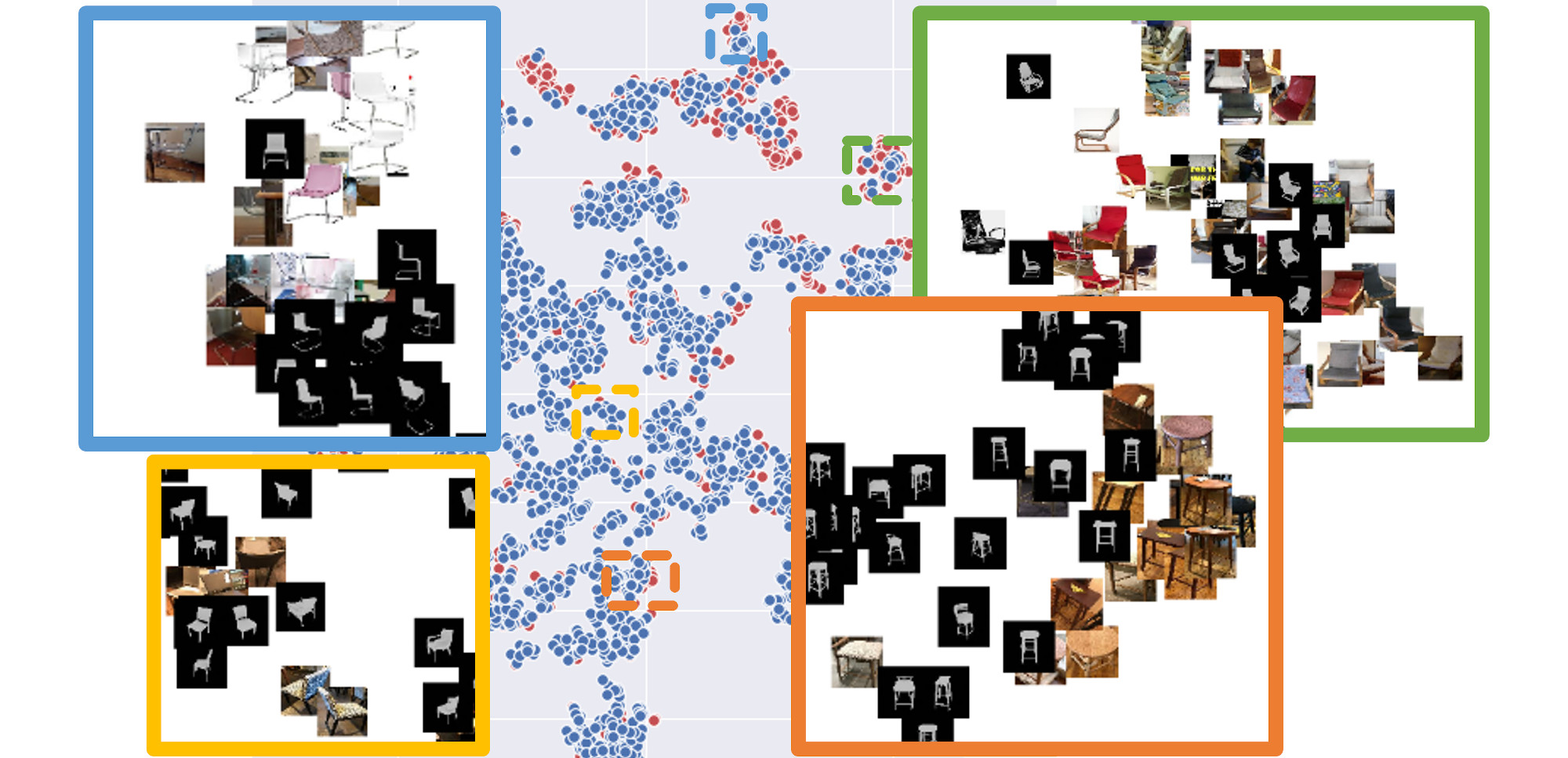} \\
    \caption{t-SNE embedding of \OURS{} for the chair class. 
    Red points correspond to images, and blue to shapes.
    }
    \label{fig:cadmask_tsne_2}
\end{figure}

\begin{figure}
    \centering
    \includegraphics[width=0.9\textwidth]{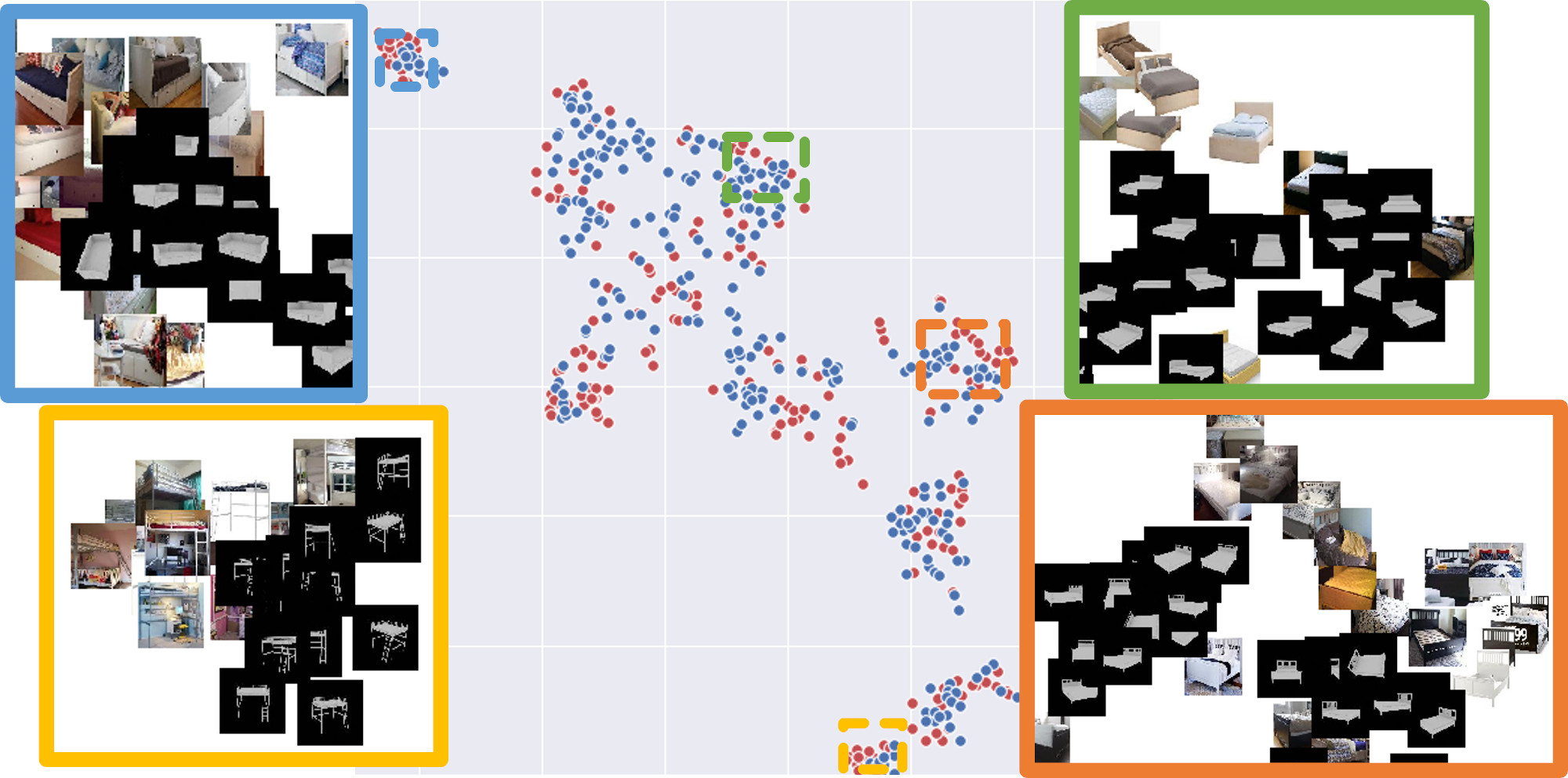} \\
    \vspace{0.1cm}
    \includegraphics[width=0.7\textwidth]{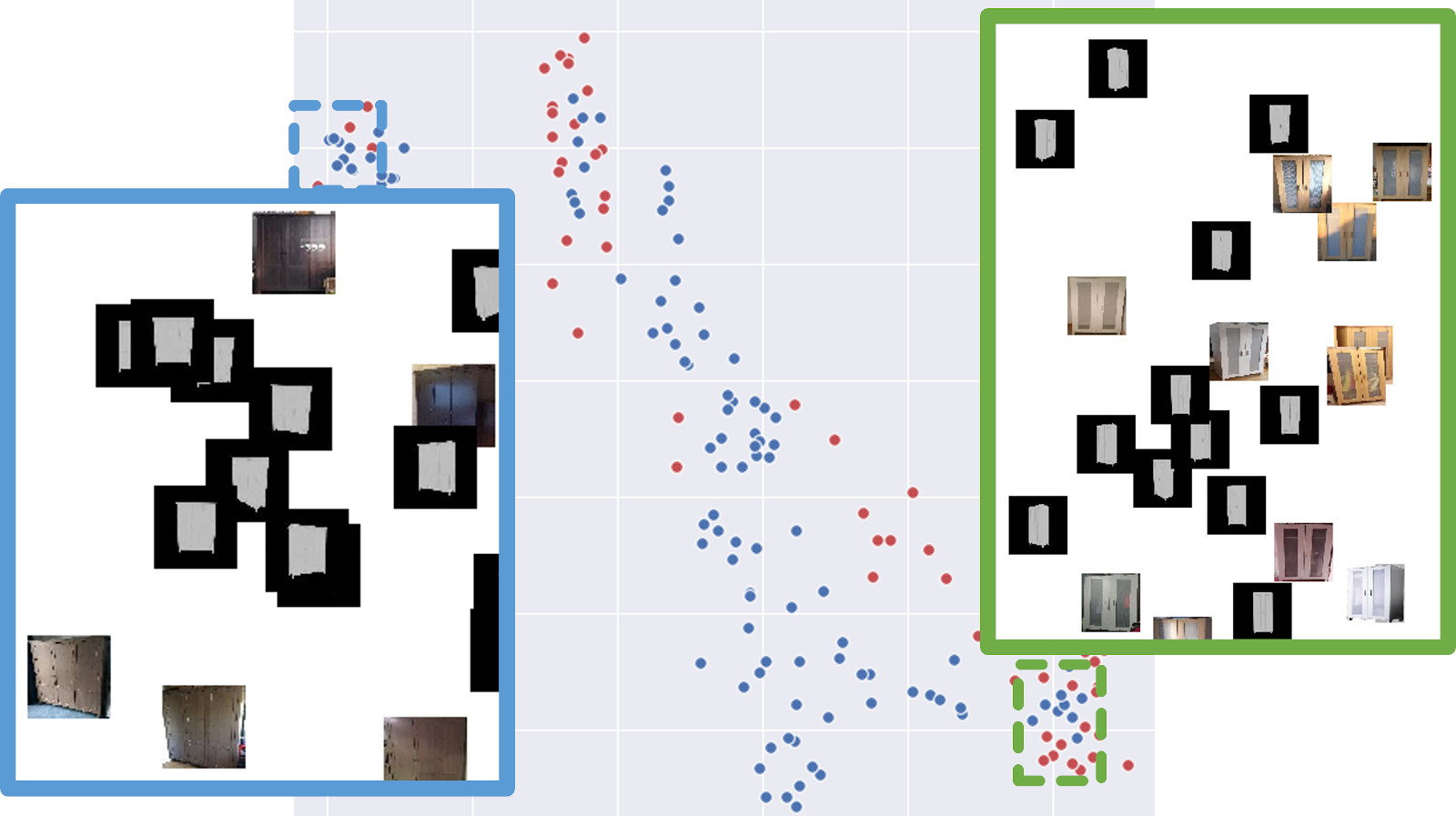} \\
    \vspace{0.1cm}
    \includegraphics[width=0.9\textwidth]{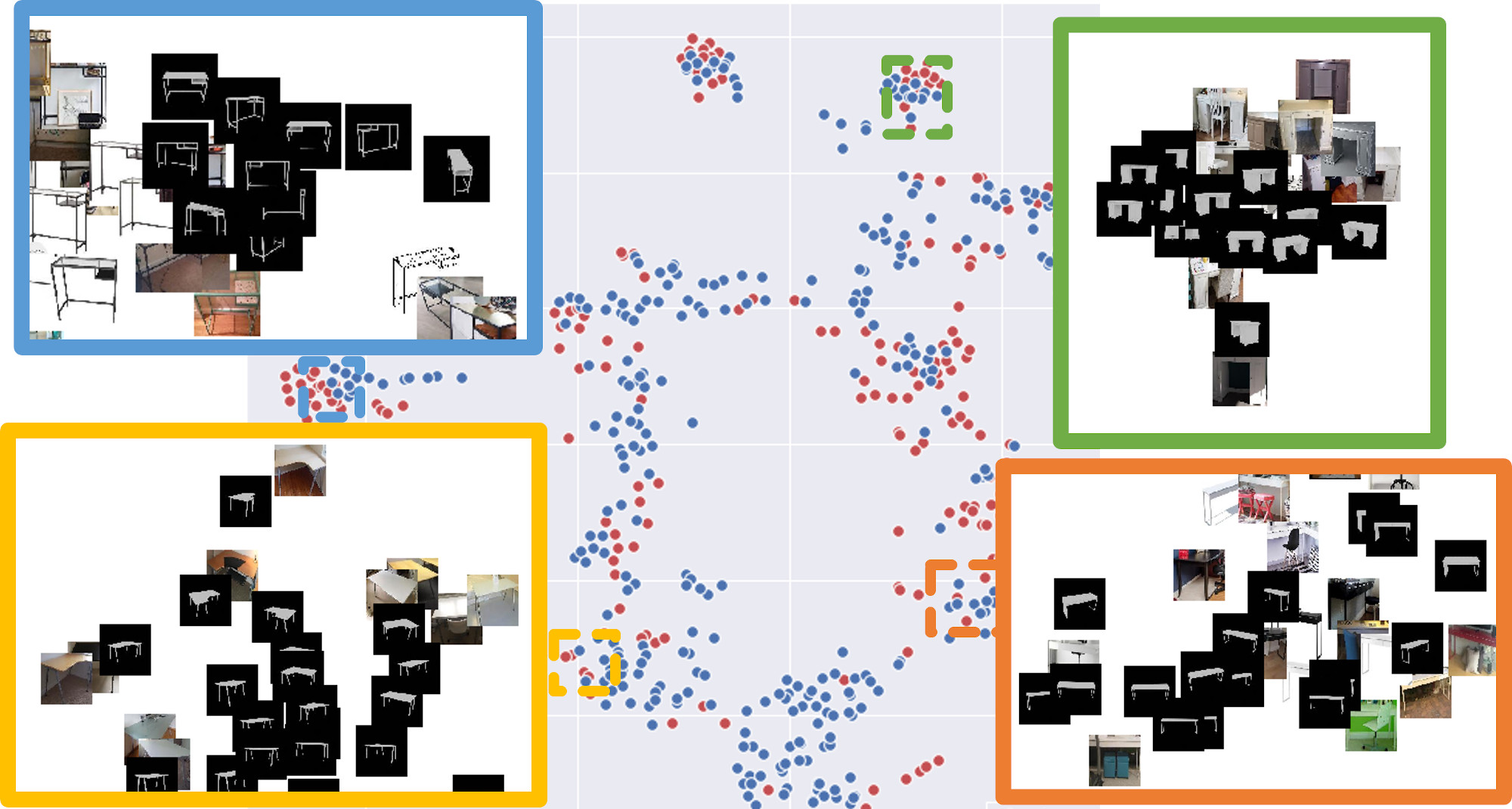} \\
    \caption{t-SNE embeddings of \OURS{} for the bed (top), wardrobe (middle) and desk (bottom) classes. 
    Red points correspond to images, and blue to shapes.
    }
    \label{fig:cadmask_tsne_0}
\end{figure}

\begin{figure}
    \centering
    \includegraphics[width=\textwidth]{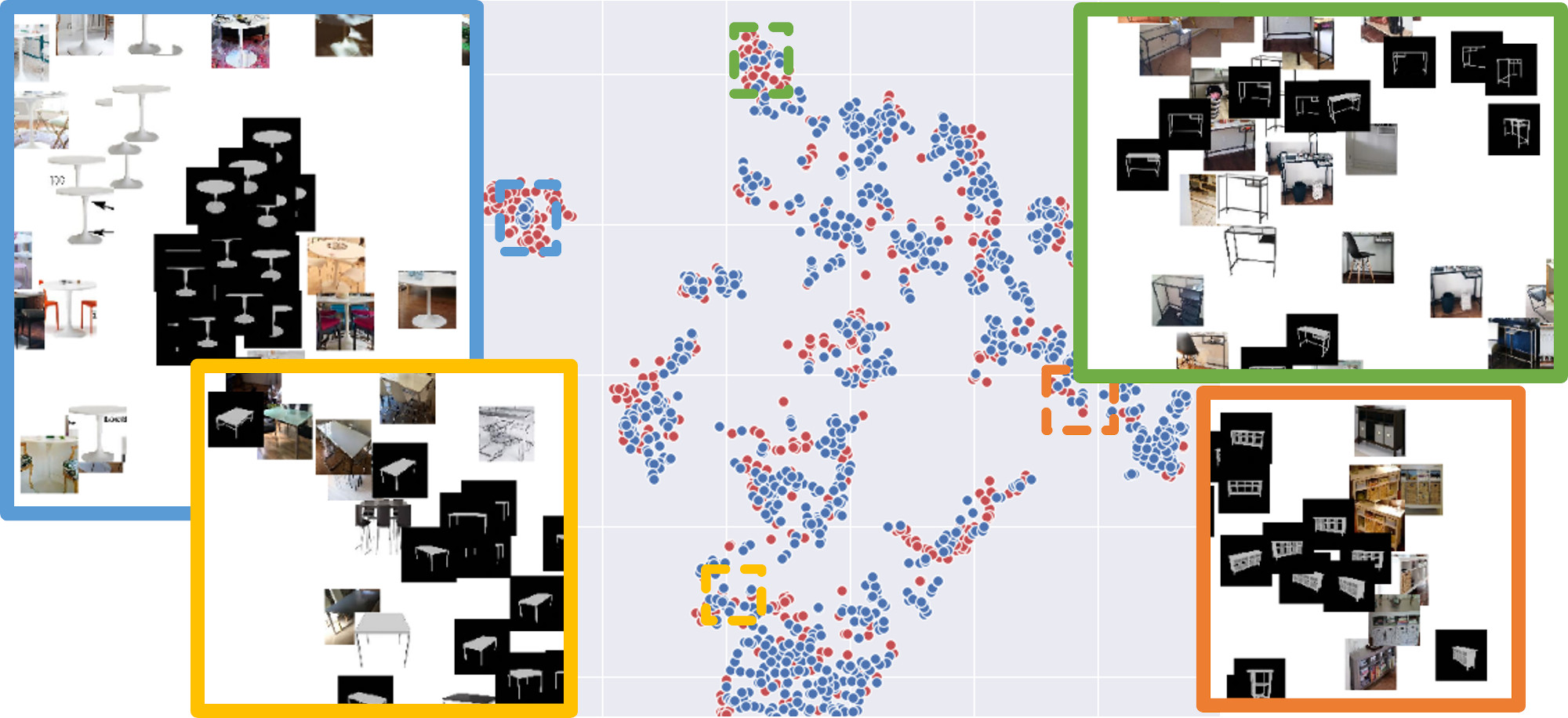} \\
    \vspace{0.1cm}
    \centering
    \includegraphics[width=0.85\textwidth]{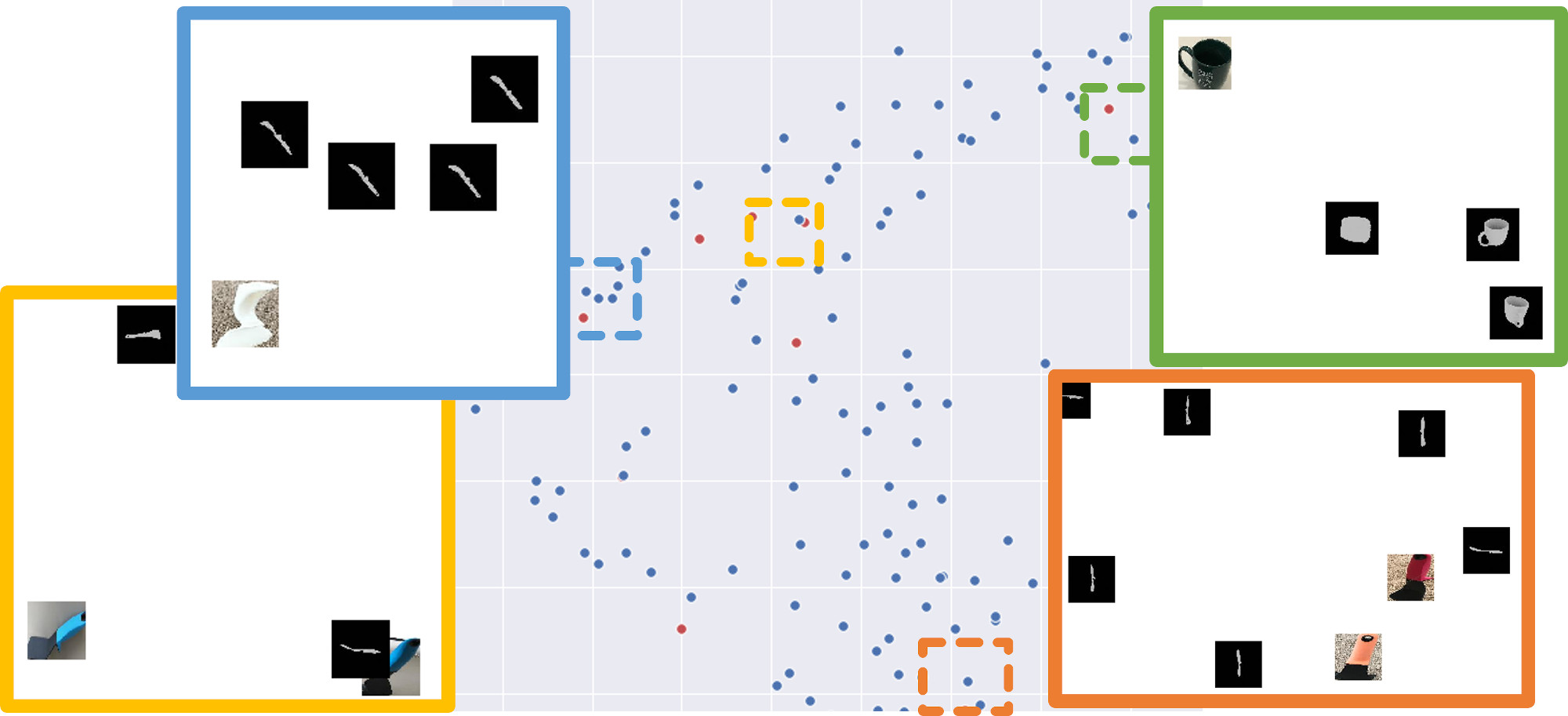} \\
    \vspace{0.1cm}
    \includegraphics[width=0.85\textwidth]{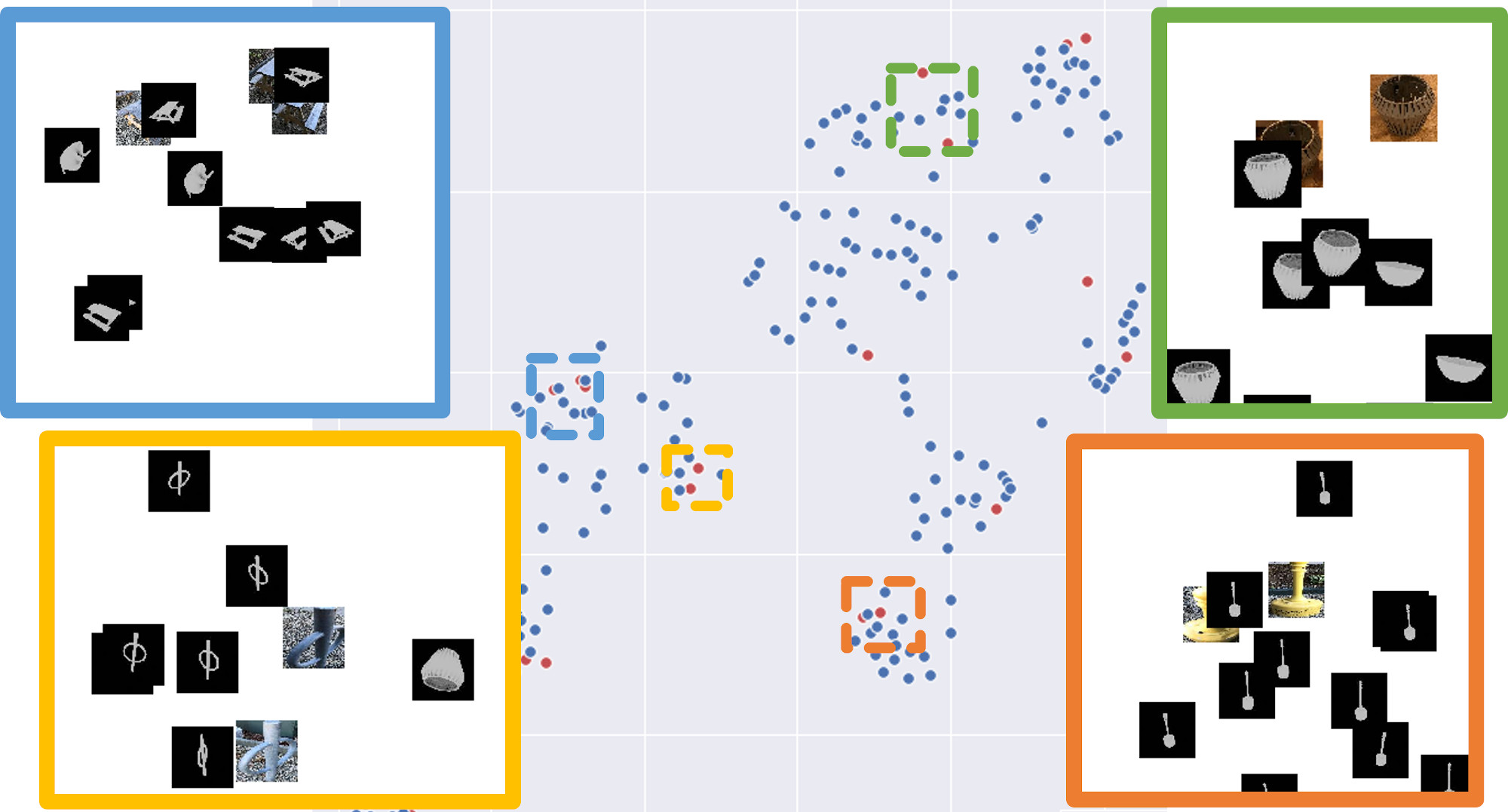} \\
    \caption{t-SNE embeddings of \OURS{} for the table (top), tool (middle) and misc (bottom) classes. 
    Red points correspond to images, and blue to shapes.
    }
    \label{fig:cadmask_tsne_1}
\end{figure}

Figures \ref{fig:cadmask_tsne_2}, \ref{fig:cadmask_tsne_0}, \ref{fig:cadmask_tsne_1} show the t-SNE visualizations of the image-shape embedding spaces for the bed, wardrobe, desk, table, tool, misc, and chair classes.

\clearpage

\end{document}